\documentclass[11pt]{article}
\usepackage[margin=1in]{geometry}
\usepackage{graphicx}
\usepackage{authblk}
\usepackage{caption}
\usepackage{subfigure}
\usepackage{subcaption}
\usepackage{float}
\usepackage{url}
\usepackage{hyperref}
\usepackage[numbers]{natbib}
\usepackage{doi}
\usepackage{multirow}
\usepackage[dvipsnames]{xcolor}
\usepackage{booktabs}
\usepackage{fancyvrb}
\usepackage{tabularx}
\usepackage{orcidlink} 
\usepackage{array}

\title{\textbf{Kvasir-VQA-x1: \\A Multimodal Dataset for Medical Reasoning and Robust MedVQA in Gastrointestinal Endoscopy}}
\author[1,2]{Sushant Gautam\,\orcidlink{0000-0001-9232-2661}}
\author[3]{Michael A. Riegler\,\orcidlink{0000-0002-3153-2064}}
\author[1,2]{Pål Halvorsen\,\orcidlink{0000-0003-2073-7029}}

\affil[1]{Simula Metropolitan Center for Digital Engineering (SimulaMet), Norway}
\affil[2]{Oslo Metropolitan University (OsloMet), Norway}
\affil[3]{Simula Research Laboratory, Norway}

\date{}

\begin{document}
\maketitle
\vspace{-3em}

\begin{abstract}
Medical Visual Question Answering (MedVQA) is a promising field for developing clinical decision support systems, yet progress is often limited by the available datasets, which can lack clinical complexity and visual diversity. To address these gaps, we introduce Kvasir-VQA-x1, a new, large-scale dataset for gastrointestinal (GI) endoscopy. Our work significantly expands upon the original Kvasir-VQA by incorporating 159,549 new question-answer pairs that are designed to test deeper clinical reasoning. We developed a systematic method using large language models to generate these questions, which are stratified by complexity to better assess a model's inference capabilities. To ensure our dataset prepares models for real-world clinical scenarios, we have also introduced a variety of visual augmentations that mimic common imaging artifacts. The dataset is structured to support two main evaluation tracks: one for standard VQA performance and another to test model robustness against these visual perturbations. By providing a more challenging and clinically relevant benchmark, Kvasir-VQA-x1 aims to accelerate the development of more reliable and effective multimodal AI systems for use in clinical settings. The dataset is fully accessible and adheres to FAIR data principles, making it a valuable resource for the wider research community. 

\noindent\textbf{Code and data:} \\\href{https://github.com/Simula/Kvasir-VQA-x1}{github.com/Simula/Kvasir-VQA-x1} and \href{https://huggingface.co/datasets/SimulaMet/Kvasir-VQA-x1}{huggingface.co/datasets/SimulaMet/Kvasir-VQA-x1}
\end{abstract}


\section{Background \& Summary}
Medical AI is witnessing a transformative shift—from pattern recognition to context-aware reasoning—driven by advances in multimodal learning. Within this landscape, Medical Visual Question Answering (MedVQA) has emerged as a compelling benchmark for evaluating real-world capabilities of vision-language models.

\subsection{The Promise and Challenge of Medical Visual Question Answering}

Medical Visual Question Answering is an emergent interdisciplinary field at the intersection of artificial intelligence, computer vision, and medicine~\cite{Lin2023Sep, VQARAD, Slake, Hartsock2024Nov}. It aims to develop systems that interpret medical images and respond to clinically pertinent questions posed in natural language~\cite{Hartsock2024Nov}. This capability holds transformative potential for clinical decision support, offering avenues to enhance diagnostic accuracy, reduce clinician workload, improve patient comprehension, and enable more equitable access to medical expertise through automated assistance in telemedicine and low-resource settings~\cite{Bazi2023Mar, Wang2023Dec,Lin2023Sep,Hartsock2024Nov}.

Unlike general-domain visual question answering (VQA), MedVQA poses unique challenges due to the complexity of medical images and the depth of domain-specific knowledge required to answer clinical questions~\cite{Liang2024Sep}. Visual cues in medical imaging are often subtle and entangled with artifacts, requiring expert-level reasoning~\cite{Singhal2025Mar}. Moreover, clinical questions may demand multi-step inference or integration of prior medical knowledge, i.e., tasks that go beyond simple pattern recognition or factual recall~\cite{Chen2024Oct,Gu2024Apr,Hu2023Jul}. As such, MedVQA is increasingly viewed as a frontier task for evaluating both vision-language reasoning and real-world applicability of multimodal AI systems in high-stakes environments~\cite{Hartsock2024Nov}.

\subsection{The Paradigm Shift Towards Generative Models}

Early MedVQA systems were predominantly discriminative, selecting answers from predefined candidate sets~\cite{Zhan2020Oct,Yu2025Apr}. While effective for constrained tasks, such systems fall short when confronted with the open-ended, nuanced questions encountered in clinical practice~\cite{Guo2025Mar,Lin2023Sep}. This limitation has driven a paradigm shift toward generative models, enabled by recent breakthroughs in Large Language Models (LLMs) and Vision--Language Models (VLMs)~\cite{Dong2025Mar,Qwen2.5-VL,Flamingo,laava,gpt4}. State-of-the-art systems such as Med-Flamingo, LLaVA-Med, MedGemma, and Qwen2.5-VL exemplify this trend, combining advanced image encoders with powerful, instruction-tuned decoders capable of producing rich, context-sensitive responses~\cite{Moor2023Jul, Li2023Jun,Qwen2.5-VL,MedFlamingo,MedGemma}. These models signal a movement towards AI systems that can engage in more natural and human-like diagnostic reasoning~\cite{Moor2023Jul, Li2023Jun}.

Yet, the promise of generative MedVQA systems is hindered by a lack of suitably complex, domain-aligned datasets for training and evaluation~\cite{PMCVQA}. Most existing benchmarks, including VQA-RAD~\cite{VQARAD}, SLAKE~\cite{VQARAD}, and PMC-VQA~\cite{PMCVQA}, suffer from small scale, limited question diversity, or over-representation of specific modalities (e.g., radiology). Many question-answer (QA) pairs are simple, fact-based, or automatically generated, which risk introducing noise and fail to promote advanced clinical reasoning~\cite{PMCVQA}. Moreover, traditional evaluation metrics (e.g., BLEU and ROUGE) often fall short in capturing the correctness or clinical utility of generative outputs, highlighting a need for new benchmarks and assessment methods rooted in real-world use cases and clinician feedback~\cite{Abbasian2024Mar,GREEN,BLEURT, BERTScore}.

\subsection{GI Endoscopy: A Unique Frontier for VQA}

\begingroup
\hyphenpenalty=10000 
\exhyphenpenalty=10000

\begin{table}[!ht]
\centering
\scriptsize
\caption{Comparison of Existing Medical VQA Datasets with Relevance to Gastrointestinal (GI) Applications}
\begin{tabular}{|p{1.8cm}|c|p{1.8cm}|p{3cm}|p{2.4cm}|p{3.8cm}|}
\hline
\textbf{Dataset} & \textbf{Year} & \textbf{Primary Focus / Modality} & \textbf{Size (Images / QA Pairs)} & \textbf{Key GI-Relevant Question Types} & \textbf{Notable Limitations for Advanced GI VQA} \\
\hline
Kvasir-VQA \cite{kvasirvqa} & 2024 & GI Endoscopy & 6,500 / $\sim$58,849 & Yes/No, choice, color, location, count & Potential for limited QA complexity for deep reasoning; scope for more conditions \\
\hline
MedVQA-GI \cite{MedVQA-GI2023} & 2023 & GI Endoscopy (Colonoscopy) & $\sim$4,000 / Multiple per image & Polyp count/location, instrument ID, color, artifacts & Incomplete expert validation; some subjective answers; limited question diversity planned for expansion \\
\hline
PMC-VQA \cite{PMCVQA} & 2023 & General Medical (80\% Radiology) & 149k / 227k & Modality, organ, abnormality (general) & Limited GI-specific focus; QA generation process risks \\
\hline
SLAKE \cite{Slake} & 2021 & Radiology (CT, MRI, X-ray) & 642 / $\sim$14k & Organ, position, abnormality (general) & Small scale; not GI-specific \\
\hline
VQA-RAD \cite{VQARAD} & 2018 & Radiology & 315 / $\sim$3,500 & Abnormality, modality, organ (general) & Very small scale; simple QA; not GI-specific \\
\hline
MedFrameQA \cite{MedFrameQA} & 2024 & Multi-image Medical Reasoning & 9,237 frames / 2,851 & Multi-image comparative reasoning & Not specifically GI-focused; tests general MLLM reasoning \\
\hline
\end{tabular}
\label{tab:gi-vqa-datasets}
\end{table}

\endgroup

Gastrointestinal (GI) endoscopy is a critical diagnostic and interventional tool in medicine, generating large volumes of high-resolution images that are rich in clinical content but visually complex~\cite{HyperKvasir}. These images frequently contain artifacts such as specular reflections, motion blur, and variable lighting, making them a challenging modality for automated interpretation~\cite{Ali2019Apr,KvasirInstrument}. Despite this, the GI domain has received relatively limited attention in VQA research compared to radiology or pathology~\cite{kvasirvqa}. Table \ref{tab:gi-vqa-datasets} provides a brief comparison of existing GI-focused VQA datasets.

Notable GI-specific resources such as HyperKvasir~\cite{HyperKvasir}, Kvasir-Instrument~\cite{KvasirInstrument}, and Kvasir-VQA~\cite{kvasirvqa} have laid important groundwork, but they often feature QA pairs centered on simple tasks, such as identifying the presence of a polyp or recognizing a tool, and thus do not fully capture the reasoning depth required for advanced clinical understanding. Similarly, while MedVQA-GI (2023) introduces diverse QA types for colonoscopy, limitations in expert validation and linguistic diversity constrain its utility for training robust generative models~\cite{hicks2023overview}.

\subsection{Towards Robust, Reasoning-Centric Evaluation}

To drive progress in this domain, we introduce \textbf{Kvasir-VQA-x1}, a significantly expanded and meticulously curated dataset designed to benchmark reasoning-intensive visual question answering in GI endoscopy. Building upon the rich visual foundations of HyperKvasir~\cite{HyperKvasir}, Kvasir-Instrument~\cite{KvasirInstrument} and Kvasir-VQA~\cite{kvasirvqa}, Kvasir-VQA-x1 features a substantially augmented corpus of question--answer pairs that target higher-order reasoning, multi-faceted clinical knowledge, and linguistic variability~\cite{qwen3technicalreport,DeBERTaV3}. Augmented questions were created using a structured, LLM-assisted pipeline followed by expert validation to ensure medical realism, linguistic fluency, and answerability from the associated image content.

Kvasir-VQA-x1 further incorporates \textit{image-level perturbations}, including occlusions, contrast shifts, and blur, to support robustness testing under varied clinical imaging conditions~\cite{Albumentations,SimCLR}. Each QA pair is additionally annotated with \textit{quantitative complexity scores}, capturing both visual and linguistic difficulty, thereby enabling nuanced stratification of model performance. This complexity-aware structuring aligns with current calls in the AI community for richer evaluation benchmarks that reflect real-world diagnostic challenges rather than artificial constraints.

Moreover, the dataset is structured into \textit{dual evaluation tracks}:
\begin{itemize}
    \item \textbf{Track 1:} Evaluates core MedVQA performance on standard images and QA pairs.
    \item \textbf{Track 2:} Assesses generalization using perturbed images.
\end{itemize}
This dual-track framework facilitates transparent comparison across models while surfacing failure modes that traditional benchmarks may obscure.

\subsection{Contribution and Outlook}

Kvasir-VQA-x1 represents a new benchmark tailored for the next generation of MedVQA systems. It addresses core limitations of prior datasets through its combination of:
\begin{itemize}
    \item \textbf{Clinical depth:} capturing nuanced reasoning in GI endoscopy,
    \item \textbf{Linguistic diversity:} enabling evaluation of generative language fluency,
    \item \textbf{Visual robustness:} stress-testing perception under real-world conditions,
    \item \textbf{Complexity scoring:} supporting layered evaluation and insight into model weaknesses.
\end{itemize}

We anticipate that this resource will serve both as a robust benchmark for state-of-the-art VLMs like MedGemma and Qwen2.5-VL, and as a catalyst for methodological innovation in medical multimodal AI. Furthermore, the dataset is prepared and released in accordance with the FAIR Data Principles, ensuring that it is Findable, Accessible, Interoperable, and Reusable. This makes Kvasir-VQA-x1 a valuable community asset for advancing clinically relevant and trustworthy AI in gastroenterology and beyond.

\section{Methods}

\subsection{Data Acquisition}
The Kvasir-VQA-x1 dataset builds upon the original Kvasir-VQA dataset~\cite{kvasirvqa}, which is itself derived from two public medical image resources: HyperKvasir~\cite{HyperKvasir} and Kvasir-Instrument~\cite{KvasirInstrument}. These datasets contain high-resolution images from gastrointestinal (GI) endoscopy procedures and are widely used in medical image analysis research. 

Kvasir-VQA was constructed by pairing 6,500 images from these sources with 58,849 annotated question-answer (QA) pairs. Medical professionals contributed to the annotation process, ensuring clinical relevance and quality. The original QA pairs span six distinct categories: Yes/No, single-choice, multiple-choice, color-related, location-related, and numerical count questions~\cite{kvasirvqa}.

To expand on this foundation and support more advanced research, we created Kvasir-VQA-x1 by applying structured augmentation strategies to both the language and visual components of the dataset. This involved generating new complex QA pairs and augmenting original images to increase diversity and robustness.

\subsection{Input Data}
The base dataset, Kvasir-VQA~\cite{kvasirvqa}, consists of 6,500 GI endoscopic images sourced from the HyperKvasir~\cite{HyperKvasir} and Kvasir-Instrument~\cite{KvasirInstrument} datasets with 58,849 QA pairs, labeled with medical expert input.
Each image is associated with multiple QA pairs that address GI findings, abnormalities, anatomical landmarks, and the presence of medical instruments. These QA pairs tend to be concise and fact-based.

The Kvasir-VQA-x1 dataset enhances the original content by introducing newly generated complex question-answer pairs through linguistic rephrasing and merging. It also includes augmented visual variants of the original 6,500 images using weak transformation pipelines such as affine transformations, cropping, and rotation. Each entry in the dataset stores the original or augmented image, the newly formulated question, a naturalized answer, a complexity score, and the original question-answer pair(s) from which it was derived.

\subsection{Processing}
To generate Kvasir-VQA-x1, we introduced two major enhancements:

\paragraph{Generation of Complex Question-Answer Pairs}
To promote reasoning beyond simple recall, we employed a structured pipeline:
\begin{itemize}
    \item \textbf{QA Grouping}: All QA pairs for a given image were grouped. A predefined list of trivial questions (e.g., "Is this finding easy to detect?", "none") was excluded to preserve quality.
    \item \textbf{Combinatorial Sampling}: We sampled sets of 1, 2, or 3 distinct QA pairs per image, balancing linguistic complexity and sample count.
    \item \textbf{Prompt Engineering}:
    \begin{itemize}
        \item \textit{Answer Naturalization}: Short answers were transformed into fluent, medically appropriate natural language.
        \item \textit{Question Merging}: When multiple QA pairs were sampled, questions were merged into a single coherent prompt requiring multi-step reasoning.
        \item \textit{Strict Formatting}: All outputs followed strict formatting rules, including JSON-encodable structure, and avoided copying raw answers.
    \end{itemize}
\end{itemize}

For question merging and answer naturalization, we used a locally hosted inference server for Qwen3-30B-A3B~\cite{qwen3technicalreport} language model. This model was chosen for its high performance, low inference cost via mixture-of-experts (MoE), strong reasoning ability, and efficient local deployment.

Each new QA pair consists of an image (either original or augmented), a complex question, a naturalized answer, the JSON-encoded original QA(s), and an integer complexity score ranging from 1 to 3, which reflects the number of original questions that have been combined. This approach enhances linguistic diversity, promotes the generation of more realistic medical language, and encourages information synthesis. The inclusion of a complexity score further enables stratified or curriculum-based training and evaluation, supporting more nuanced model development and assessment. Table \ref{tab:example} illustrates an example image from the Kvasir-VQA dataset along with newly generated question–answer pairs of varying complexity levels, where three representative samples from each category are shown.

\paragraph{Weak Image Augmentation for Enhanced Robustness}
To account for minor variations in imaging (e.g., due to camera angle or lighting), we generated 10 weakly augmented versions for each original image using:
\begin{itemize}
    \item \texttt{RandomResizedCrop} (scale: 0.9–1.0)
    \item \texttt{RandomRotation} (±10 degrees)
    \item \texttt{RandomAffine} (translation up to 10\%)
    \item \texttt{ColorJitter} (brightness and contrast: 0.8–1.2)
\end{itemize}
All augmentations used bicubic interpolation and appropriate padding.
During dataset construction, each QA pair was paired with either the original image ($\sim$23\% of cases) or an augmented version ($\sim$77\%) to simulate realistic variability. The released dataset only includes the original images, associated QA pairs, and metadata. Augmented images are not included in the dataset but can be generated using our provided augmentation script. We provide scripts to reproduce the exact augmented images used during generation, along with corresponding train/test splits.
These controlled augmentations introduce visual variance while preserving semantic meaning. The probabilistic mix ensures models are exposed to both clean and slightly perturbed data, enhancing generalization to real-world clinical settings.

\paragraph{Dataset Statistics and Splits}
The final Kvasir-VQA-x1 dataset comprises 159,549 question-answer (QA) pairs. Each entry references an original image and includes a complex question, naturalized answer, original QA metadata, and a complexity score.

We release the dataset with only the original images (no augmentations) and provide scripts to generate weakly augmented versions of each image. We define two \textbf{evaluation settings}:

\begin{enumerate}
    \item \textbf{Original Setting} — QA pairs referencing only the original, unaltered images.
    \item \textbf{Transformed Setting} — QA pairs referencing weakly augmented images, generated using the provided scripts.
\end{enumerate}

These settings enable flexible experimentation, including training on clean data and testing robustness under visual perturbations. Each image-question pair is assigned to either the training or test set, ensuring that no identical QA instance appears in both. This setup supports meaningful generalization testing and is well-suited for training deep learning models and conducting reliable benchmark evaluations in the medical VQA domain.
\begin{table}[htbp]
\centering
\caption{
An example image and its associated question–answer pairs, stratified by complexity. Three samples are shown from each complexity category.
}
\vspace{2mm}
\includegraphics[width=0.4\linewidth]{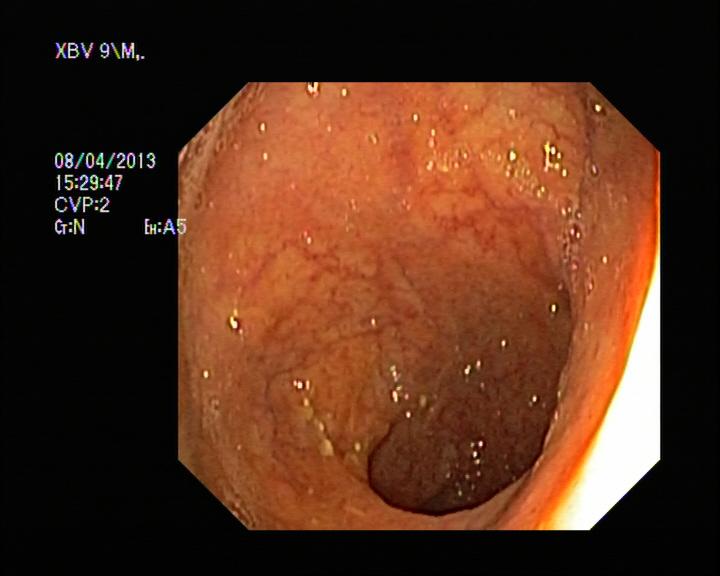}
\vspace{2mm}

\resizebox{\textwidth}{!}{%
\begin{tabular}{@{}c >{\raggedright\arraybackslash}p{6cm} >{\raggedright\arraybackslash}p{6.2cm} >{\raggedright\arraybackslash}p{4cm}@{}}
\toprule
\textbf{Complexity} & \textbf{Question} & \textbf{Answer} & \textbf{Question Class} \\
\midrule
1 & Is there any remaining polyp tissue present? & No polyp remnants identified & polyp\_removal\_status \\
1 & Are there any abnormalities present in the gastrointestinal image? & evidence of ulcerative colitis & abnormality\_presence \\
1 & Which anatomical landmark is visible in the image? & No identifiable anatomical landmark present & landmark\_location \\
2 & Are there any green or black box artifacts present, and what are the colors of the abnormality? & No green or black box artifacts observed, with abnormalities appearing pink and red & box\_artifact\_presence, abnormality\_color \\
2 & What type of polyp is present and how many findings are visible? & No polypoidal lesions identified, with a single abnormal finding noted. & polyp\_type, finding\_count \\
2 & What procedure is depicted in the image and what colors are associated with the abnormality? & Evidence of colonoscopy findings with pink and red mucosal lesions & procedure\_type, abnormality\_color \\
3 & Are there any abnormalities, anatomical landmarks, or polyps visible in the image? & No significant abnormalities or polyps were identified, and no anatomical landmarks were noted. & abnormality\_presence, landmark\_presence, polyp\_size \\
3 & Are there any anatomical landmarks visible, what procedure was performed, and is there an instrument present in the image? & No anatomical landmarks are visible, the image is from a colonoscopy, and no instrument is present. & landmark\_presence, procedure\_type, instrument\_location \\
3 & Are there any anatomical landmarks visible, what type of polyps are present, and what colors are the observed abnormalities? & No anatomical landmarks identified, no polyps observed, and multiple abnormalities with pink and red coloration. & landmark\_presence, polyp\_type, abnormality\_color \\
\bottomrule
\end{tabular}
}
\label{tab:example}
\end{table}

\subsection{Model Fine-Tuning and Evaluation Strategy}

This section outlines the strategic approach for fine-tuning vision-language models and the comprehensive evaluation framework employed. The objective is to enhance model performance on GI endoscopy image analysis and question answering, while also rigorously assessing their robustness and generalization capabilities.

\textbf{Model Fine-Tuning}

We fine-tune two prominent vision-language models, MedGemma~\cite{MedGemma} and Qwen2.5-VL~\cite{Qwen2.5-VL}, to adapt them to our specific dataset of GI endoscopy images. Fine-tuning aims to leverage the pre-trained knowledge of these models and specialize them for the nuances of medical image understanding and clinical question answering. The suffix ``\texttt{-ft}'' is appended to denote fine-tuned variants (e.g., \texttt{medgemma-ft}).

The fine-tuning process utilizes LoRA (Low-Rank Adaptation). LoRA is chosen because it allows for efficient adaptation of large pre-trained models by injecting trainable low-rank matrices into the transformer layers, primarily targeting the language model components~\cite{lora,peft-paper,SafaviNaini2024Aug}. This approach significantly reduces the number of trainable parameters, making fine-tuning computationally less demanding while maintaining strong performance~\cite{Lian2024Jan}.

For Qwen2.5-VL, we explore two distinct fine-tuning variants to investigate the impact of data augmentation:
\begin{itemize}
    \item \texttt{Qwen2.5-VL-ft-XXXX}: This variant is fine-tuned using the original image set. This provides a baseline understanding of how the model performs when exposed only to the raw data.
    \item \texttt{Qwen2.5-VL-ft-Trans-XXXX}: This variant is fine-tuned using a transformed (augmented) image set. The inclusion of augmented images is designed to improve the model's robustness and generalization by exposing it to a wider variety of visual perturbations, thereby making it less susceptible to minor variations in input.
\end{itemize}

\textbf{Evaluation Strategy}

A robust evaluation strategy is critical to comprehensively assess the performance of the fine-tuned models~\cite{SafaviNaini2024Aug}. This involves benchmarking against established baselines, evaluating across different datasets, and employing a diverse set of metrics~\cite{Islam2024Jun}.

\clearpage

\textbf{Benchmarking Baselines}

Fine-tuned models are rigorously benchmarked against their base versions: \texttt{gemma3}, \texttt{medgemma}, and \texttt{Qwen2.5-VL} (non-fine-tuned). These baselines serve as crucial reference points, demonstrating the performance improvements attributable to the fine-tuning process. Where applicable, these baselines represent publicly available checkpoints, facilitating reproducibility and comparative analysis with external research.

\textbf{Evaluation Datasets}

Performance is evaluated using two distinct datasets to assess both primary performance and robustness:
\begin{itemize}
    \item \textbf{Normal}: This dataset comprises the original GI endoscopy images. It is used to gauge the models' primary performance on the core task of understanding and answering questions based on untransformed clinical images.
    \item \textbf{Transformed}: This dataset consists of weakly augmented versions of the GI endoscopy images. Evaluating on this set assesses the models' robustness under visual perturbations, indicating their ability to generalize and maintain performance when faced with minor visual variations, which are common in real-world clinical settings.
\end{itemize}

\textbf{Performance Metrics}

Models are assessed using a comprehensive suite of standard VQA and natural language processing (NLP) metrics~\cite{Gao2025}, chosen to capture various facets of response quality:
\begin{itemize}
    \item \textbf{ROUGE-1, ROUGE-2, ROUGE-L}: These metrics measure n-gram overlap and sequence similarity, providing insights into the content overlap between the model's answer and the ground truth~\cite{Rouge}.
    \item \textbf{METEOR}: This metric goes beyond simple n-gram overlap by capturing synonymy and stem similarity, offering a more nuanced assessment of semantic equivalence~\cite{Meteor}.
    \item \textbf{CHRF++}: A character-level F-score metric, particularly useful for evaluating text in morphologically rich contexts, ensuring that character-level matches are also considered~\cite{chrF}.
    \item \textbf{BLEU}: A widely used n-gram precision-based translation metric, traditionally employed in machine translation, but also valuable for evaluating the fluency and accuracy of generated text~\cite{BLEU}.
    \item \textbf{BLEURT}: A learned evaluation metric based on BERT modeling of human quality judgments. This metric aims to align more closely with human perceptions of answer quality, offering a more holistic assessment~\cite{BLEURT}.
    \item \textbf{BERT-F1}: An embedding-based similarity metric with F1 aggregation using BERT. This metric assesses the semantic similarity between the generated answer and the ground truth by leveraging contextual embeddings from BERT~\cite{BERTScore}.
\end{itemize}

\textbf{Evaluation Granularity}

To provide a detailed understanding of model capabilities, evaluation is performed at multiple granularities:
\begin{itemize}
    \item \textbf{Intermediate Checkpoint Evaluation}: Performance is evaluated across intermediate checkpoints (e.g., \texttt{medgemma-2000}, \texttt{medgemma-3000}, \texttt{medgemma-3952}). This allows for the observation of learning trends throughout the fine-tuning process and the identification of optimal performance snapshots before potential overfitting.
    \item \textbf{Overall Performance}: Aggregate scores are computed across all complexity levels and question categories to summarize general model performance and derive comparative insights into the overall effectiveness of fine-tuning.
    \item \textbf{Categorical Evaluation}: Model performance is broken down across 18 specific clinical question categories (e.g., \texttt{abnormality\_color}, \texttt{finding\_count}, \texttt{polyp\_type}, \texttt{instrument\_presence}). This fine-grained analysis helps identify strengths and weaknesses in specific clinical reasoning areas. Visualizations like Rank-Normalized Heatmaps (for relative ranking) and Radar Charts (for absolute category-wise scores) are used to illustrate these insights.
    \item \textbf{Complexity-Based Evaluation}: Questions are categorized into three levels of reasoning complexity:
    \begin{itemize}
        \item \textbf{Level 1:} Direct prompts derived from a single atomic QA pair, requiring straightforward factual recall.
        \item \textbf{Level 2:} Prompts created by merging two atomic QA pairs, demanding moderate reasoning and synthesis across related clinical cues.
        \item \textbf{Level 3:} Prompts combining three distinct QA pairs into a single question, requiring higher-order reasoning, abstraction, and cross-referencing multiple clinical aspects.
    \end{itemize}
    This tiered evaluation quantifies the models' robustness in handling increasing reasoning demands and linguistic diversity, providing insights into their ability to perform complex clinical inference.
\end{itemize}

\subsection{Automated Evaluation using an LLM-based Adjudicator}

To address the limitations of traditional n-gram-based metrics (e.g., BLEU, ROUGE) in capturing clinical accuracy and semantic correctness, we implemented a sophisticated, automated evaluation protocol using a powerful Large Language Model (LLM) as a structured adjudicator. This methodology provides a fine-grained, categorical assessment of model performance, directly aligning with the clinical reasoning aspects defined in our dataset.

The core of this evaluation is a programmatic pipeline that leverages the \texttt{Qwen/Qwen3-30B-A3B} model as an impartial medical examiner. For each prediction made by a model being tested, a detailed, structured prompt is generated. This prompt provides the adjudicator LLM with comprehensive context, including:

\begin{enumerate}
    \item \textbf{The Endoscopic Image Question:} The input question posed to the model.
    \item \textbf{The Model's Generated Response:} The answer produced by the fine-tuned model (e.g., \texttt{MedGemma-ft}, \texttt{Qwen2.5-VL-ft}).
    \item \textbf{The Ground-Truth Answer:} The correct answer from the Kvasir-VQA-x1 dataset.
    \item \textbf{Evaluation Aspects:} A list of specific clinical categories (\texttt{question\_class}) that the question pertains to (e.g., \texttt{polyp\_type}, \texttt{instrument\_presence}, \texttt{abnormality\_location}).
    \item \textbf{Ancillary Context:} The question's complexity level (1--3) and the original, atomic QA pairs from which the complex question was derived, sourced from the Kvasir-VQA dataset.
\end{enumerate}

The adjudicator LLM is instructed to act as a medical examiner grading an exam. Its task is to systematically compare the model's response against the ground-truth answer, focusing \textit{only} on the specified evaluation aspects. For each aspect, it must assign a binary score: `1` if the model's response correctly and completely addresses that specific aspect, and `0` if it is incorrect, incomplete, or fails to address it. A brief textual justification for the score is also required.

The entire process is automated using an asynchronous Python script that sends batched requests to a hosted endpoint for the \texttt{Qwen/Qwen3-30B-A3B}~\cite{qwen3technicalreport} model. The adjudicator's output is captured in a structured JSON format:

\begin{Verbatim}[samepage=true]
{
  "eval_json": {
    "polyp_type": {
      "score": 1,
      "reason": "The model correctly identified the polyp as sessile."
    },
    "instrument_presence": {
      "score": 0,
      "reason": "The model failed to mention the presence of biopsy forceps, 
               which are visible."
    }
  }
}
\end{Verbatim}

This automated, LLM-driven adjudication process yields a rich, multi-faceted evaluation. By aggregating the binary scores on a per-category basis, we can compute the categorical accuracy metrics presented in Section~4. This approach allows us to move beyond surface-level text similarity and perform a scalable, reproducible, and semantically nuanced assessment of each model's clinical reasoning capabilities across different domains of GI endoscopy.

\subsection{Training Configuration}

We fine-tuned vision--language models on the Kvasir-VQA-x1 dataset using parameter-efficient tuning with Low-Rank Adaptation (LoRA). All experiments were conducted with standardized hyperparameters, clinical instruction prompts, and multi-GPU infrastructure. Below, we outline the training details necessary for reproducibility. The training setup, including hardware specifications and core hyperparameters, are summarized in Table \ref{tab:hyperparameters} and \ref{tab:FinetuningSummary}.

\paragraph{Model Setup.} Each model was initialized from a publicly available checkpoint and adapted using LoRA, with frozen vision backbones and trainable language layers. The instruction prompt used during fine-tuning was:

\begin{quote}
\textit{``You are a medical vision-language assistant; given an endoscopic image and a clinical question that may ask about one or more findings, provide a concise, clinically accurate response addressing all parts of the question in natural-sounding medical language as if spoken by a doctor in a single sentence.''}
\end{quote}

We employed the Hugging Face \texttt{transformers}~\cite{transformers}, \texttt{PEFT}~\cite{PEFT}, and \texttt{swift}~\cite{ms-swift} toolchains with DeepSpeed ZeRO Stage 2~\cite{ZeRO} optimization.

\begin{table}[h]
\centering
\small
\begin{tabular}{ll}
\toprule
\textbf{Attribute} & \textbf{Specification} \\
\midrule
GPUs & 4--8 $\times$ A100 (80GB) \\
Precision & \texttt{bfloat16} (DeepSpeed) \\
Framework & Hugging Face, Swift, PEFT \\
Optimizer & Fused AdamW (DeepSpeed) \\
Scheduler & Linear / Cosine (model-specific) \\
Effective Batch Size & 36 (MedGemma), 32 (Qwen2.5) \\
Max Sequence Length & 1000 tokens \\
Warmup Ratio & 0.03 \\
\bottomrule
\end{tabular}
\caption{Training environment and hyperparameters.}
\label{tab:hyperparameters}
\end{table}

\paragraph{Implementation Notes.} All models used LoRA with frozen vision encoders. For all variants, LoRA targeted all projection layers (\texttt{q\_proj}, \texttt{k\_proj}, \texttt{v\_proj}, etc.). \textbf{Reproducibility.} Fixed random seeds, and released configuration files should ensure reproducibility.  Adapter weights and training logs will be made publicly available.

\section{Data Records}

The \textbf{Kvasir-VQA-x1} dataset is hosted on the Hugging Face Datasets Hub at: \url{https://huggingface.co/datasets/SimulaMet/Kvasir-VQA-x1}. It adheres to the FAIR principles—Findable, Accessible, Interoperable, and Reusable—to support reproducible research in medical multimodal AI~\cite{FAIR}.

\subsection*{Dataset Access and Exploration}

Users can interact with the dataset through:

\begin{itemize}
    \item \textbf{Web Interface}: Browse, filter, and search the dataset directly on the Hugging Face platform.
    \item \textbf{Python API}: Load the dataset using the \texttt{datasets} library:
    \begin{verbatim}
    from datasets import load_dataset
    dataset = load_dataset("SimulaMet/Kvasir-VQA-x1")
    \end{verbatim}
    \item \textbf{Command-Line Interface (CLI)}: Utilize the Hugging Face CLI for dataset operations.
\end{itemize}

\subsection*{Dataset Structure}

Each entry in the dataset comprises the following fields:

\begin{itemize}
    \item \texttt{img\_id}: Unique identifier linking to the corresponding image from the Kvasir\_VQA dataset.
    \item \texttt{complexity}: Integer score (1, 2, or 3) indicating the reasoning complexity of the question.
    \item \texttt{question}: Natural language question derived from one or more atomic QA pairs.
    \item \texttt{answer}: Clinically validated answer corresponding to the question.
    \item \texttt{original}: JSON-encoded list of the original atomic QA pairs used to generate the complex question.
    \item \texttt{question\_class}: List of clinical categories associated with the question (e.g., \texttt{polyp\_type}, \texttt{instrument\_presence}, \texttt{finding\_count}). See Table 2 in the Kvasir\_VQA paper~\cite{kvasirvqa} for different questions in the original dataset.
\end{itemize}

\subsection*{Data Splits}

The dataset is divided into two predefined splits:

\begin{itemize}
    \item \textbf{Train}: Contains samples associated with the training subset of the original images.
    \item \textbf{Test}: Contains samples associated with the testing subset, reserved for final model evaluation.
\end{itemize}

\subsection*{Dataset Statistics}

The final Kvasir-VQA-x1 dataset includes 159,549 question–answer pairs linked to 6,500 original GI endoscopy images. Each QA pair is annotated with a reasoning complexity score (Level 1–3) and associated clinical question classes. Below, we summarize key statistics.

\begin{table}[h]
\centering
\caption{Dataset distribution by complexity level and question class}
\begin{tabular}{lrrrr}
\toprule
Category / Description & Level 1 & Level 2 & Level 3 & Total \\
\midrule
\multicolumn{5}{l}{\textit{Complexity levels (overall)}} \\
Level 1 — Simple recall from a single QA pair & 54\,856 & — & — & 54\,856\,(34.4\%) \\
Level 2 — Merged reasoning across two QA pairs  & — & 52\,349 & — & 52\,349\,(32.8\%) \\
Level 3 — Multi-hop reasoning across three QA pairs & — & — & 52\,344 & 52\,344\,(32.8\%) \\
\midrule
\multicolumn{5}{l}{\textit{Question classes}} \\
abnormality\_color          & 3\,125 & 6\,127 & 9\,355 & 18\,607 \\
abnormality\_location       & 3\,112 & 6\,233 & 9\,464 & 18\,809 \\
abnormality\_presence       & 3\,099 & 6\,193 & 9\,243 & 18\,535 \\
box\_artifact\_presence      & 3\,940 & 7\,792 & 11\,744 & 23\,476 \\
finding\_count              & 3\,429 & 6\,856 & 10\,286 & 20\,571 \\
finding\_presence           & 2\,500 & 0     & 0      & 2\,500 \\
instrument\_count           & 3\,555 & 7\,019 & 10\,698 & 21\,272 \\
instrument\_location        & 3\,154 & 6\,228 & 9\,533 & 18\,915 \\
instrument\_presence        & 3\,148 & 6\,327 & 9\,402 & 18\,877 \\
landmark\_color             & 88    & 186    & 260    & 534   \\
landmark\_location          & 2\,158 & 4\,416 & 6\,560 & 13\,134 \\
landmark\_presence          & 2\,000 & 3\,965 & 6\,066 & 12\,031 \\
polyp\_count                & 3\,134 & 6\,247 & 9\,428 & 18\,809 \\
polyp\_removal\_status      & 3\,945 & 7\,993 & 11\,661 & 23\,599 \\
polyp\_size                 & 3\,258 & 6\,590 & 9\,794 & 19\,642 \\
polyp\_type                 & 3\,331 & 6\,717 & 9\,960 & 20\,008 \\
procedure\_type             & 3\,939 & 7\,923 & 11\,805 & 23\,667 \\
text\_presence              & 3\,941 & 7\,886 & 11\,773 & 23\,600 \\
\bottomrule
\end{tabular}
\label{tab:dataset-distribution}
\end{table}

We note that the uneven per-class counts in Table \ref{tab:dataset-distribution}
 largely reflect the underlying distribution and mergeability of the original Kvasir-VQA annotations: rare classes (e.g.,\ \textit{landmark color}~\texttt{landmark\_color}) simply had few atomic QA pairs to begin with, and binary presence checks (finding presence) cannot be meaningfully composed into multi-step questions. Moreover, multi-hop QA generation requires co-occurring annotations on the same image, and our question-merging step further prevents ambiguous or clinically irrelevant merges. Together, these factors naturally constrain Level 2 and 3 sample sizes for certain categories while preserving the integrity and clinical validity of the dataset.

\section{Technical Validation}
Before diving into the empirical analysis, we now shift our focus from dataset construction to how well state-of-the-art vision-language models perform on the Kvasir-VQA-x1 benchmark. Table \ref{tab:FinetuningSummary} summarizes the fine-tuning results for both models, including training duration, accuracy, and evaluation loss.

\begin{table}[h]
\centering
\small
\begin{tabular}{lccccccc}
\toprule
\textbf{Model} & \textbf{Params} & \textbf{Epochs} & \textbf{LR} & \textbf{LoRA (r/$\alpha$)} & \textbf{Time} & \textbf{Eval Acc.} & \textbf{Eval Loss} \\
\midrule
MedGemma-Transf. & 4.3B & 4 & 2e-5 & 16 / 64 & 27 h & 84.97\% & 0.4111 \\
Qwen2.5-VL-Transf. & 8.3B & 4 & 2e-5 & 16 / 64 & 30.9 h & \textbf{85.91\%} & \textbf{0.3883} \\
Qwen2.5-VL & 8.3B & 3 & 2e-5 & 16 / 64 & 23 h & 85.78\% & 0.3906 \\
\bottomrule
\end{tabular}
\caption{Fine-tuning summary table for both models. The fine-tuning evaluation loss for all models was computed on a randomly selected 1\% subset of the training data, which was held out and not used during training.
}
\label{tab:FinetuningSummary}
\end{table}

\subsection{Model Fine-Tuning and Evaluation Strategy}

This section presents the empirical findings from the fine-tuning and evaluation of the vision-language models, highlighting their performance across various metrics, clinical categories, and reasoning complexities.

\begin{table}[H]
\centering
\scriptsize
\caption{Evaluation metrics for various models on Normal and Transformed validation sets.}
\begin{tabular}{|l|c|c|c|c|c|c|c|c|}
\hline
\textbf{Model} & \textbf{ROUGE-1} & \textbf{ROUGE-2} & \textbf{ROUGE-L} & \textbf{METEOR} & \textbf{CHRF++} & \textbf{BLEU} & \textbf{BLEURT} & \textbf{BERT-F1} \\
\hline
\multicolumn{9}{|c|}{\textbf{Normal Validation Set}} \\
\hline
gemma3 & 0.159 & 0.036 & 0.131 & 0.214 & 29.167 & 0.019 & -0.723 & 0.861 \\
medgemma-4b & 0.227 & 0.064 & 0.192 & 0.251 & 31.698 & 0.037 & -0.557 & 0.878 \\
medgemma-2000 & 0.665 & 0.457 & 0.634 & 0.634 & 62.489 & 0.404 & 0.305 & 0.946 \\
medgemma-3000 & 0.673 & 0.468 & 0.643 & 0.643 & 63.591 & 0.419 & 0.325 & 0.948 \\
medgemma-3952 & 0.677 & 0.476 & 0.648 & 0.648 & 63.931 & 0.428 & 0.333 & 0.948 \\
Qwen2.5-VL & 0.230 & 0.072 & 0.193 & 0.286 & 33.910 & 0.033 & -0.531 & 0.875 \\
Q-VL-ft-3000 & 0.715 & 0.531 & 0.689 & 0.688 & 67.792 & 0.476 & 0.402 & 0.954 \\
\textbf{Q-VL-ft-3333} & \textbf{0.716} & \textbf{0.534} & \textbf{0.690} & \textbf{0.689} & \textbf{67.910} & \textbf{0.478} & \textbf{0.404} & \textbf{0.954} \\
Q-VL-ft-Trans-3000 & 0.710 & 0.526 & 0.683 & 0.681 & 67.005 & 0.463 & 0.396 & 0.954 \\
Q-VL-ft-Trans-4444 & 0.712 & 0.529 & 0.687 & 0.686 & 67.480 & 0.472 & 0.398 & 0.954 \\
\hline
\multicolumn{9}{|c|}{\textbf{Transformed Validation Set}} \\
\hline
gemma3 & 0.158 & 0.035 & 0.130 & 0.212 & 29.038 & 0.018 & -0.734 & 0.861 \\
medgemma-4b & 0.227 & 0.063 & 0.192 & 0.250 & 31.645 & 0.037 & -0.555 & 0.878 \\
medgemma-2000 & 0.664 & 0.456 & 0.633 & 0.633 & 62.617 & 0.404 & 0.308 & 0.946 \\
medgemma-3000 & 0.673 & 0.469 & 0.644 & 0.644 & 63.659 & 0.420 & 0.326 & 0.948 \\
medgemma-3952 & 0.676 & 0.472 & 0.646 & 0.647 & 63.928 & 0.426 & 0.329 & 0.948 \\
Qwen2.5-VL & 0.229 & 0.071 & 0.192 & 0.284 & 33.835 & 0.032 & -0.529 & 0.875 \\
Q-VL-ft-3000 & 0.707 & 0.522 & 0.681 & 0.681 & 67.132 & 0.467 & 0.388 & 0.953 \\
Q-VL-ft-3333 & 0.708 & 0.524 & 0.682 & 0.682 & 67.150 & 0.467 & 0.390 & 0.953 \\
Q-VL-ft-Trans-3000 & 0.709 & 0.524 & 0.683 & 0.681 & 66.891 & 0.461 & 0.393 & 0.953 \\
Q-VL-ft-Trans-4444 & 0.711 & 0.527 & 0.685 & 0.685 & 67.413 & 0.470 & 0.396 & 0.954 \\
\hline
\end{tabular}
\label{tab:evaluation_metrics}
\end{table}

The overall performance, as measured by standard VQA and NLP metrics, is summarized in Table \ref{tab:evaluation_metrics}. These aggregate scores provide a comprehensive overview of the models' general capabilities.

\textbf{Categorical Performance} \\
The evaluation broke down model performance across 18 distinct clinical question categories.

\begin{figure}[htpb]
    \centering
    \includegraphics[width=0.8\linewidth]{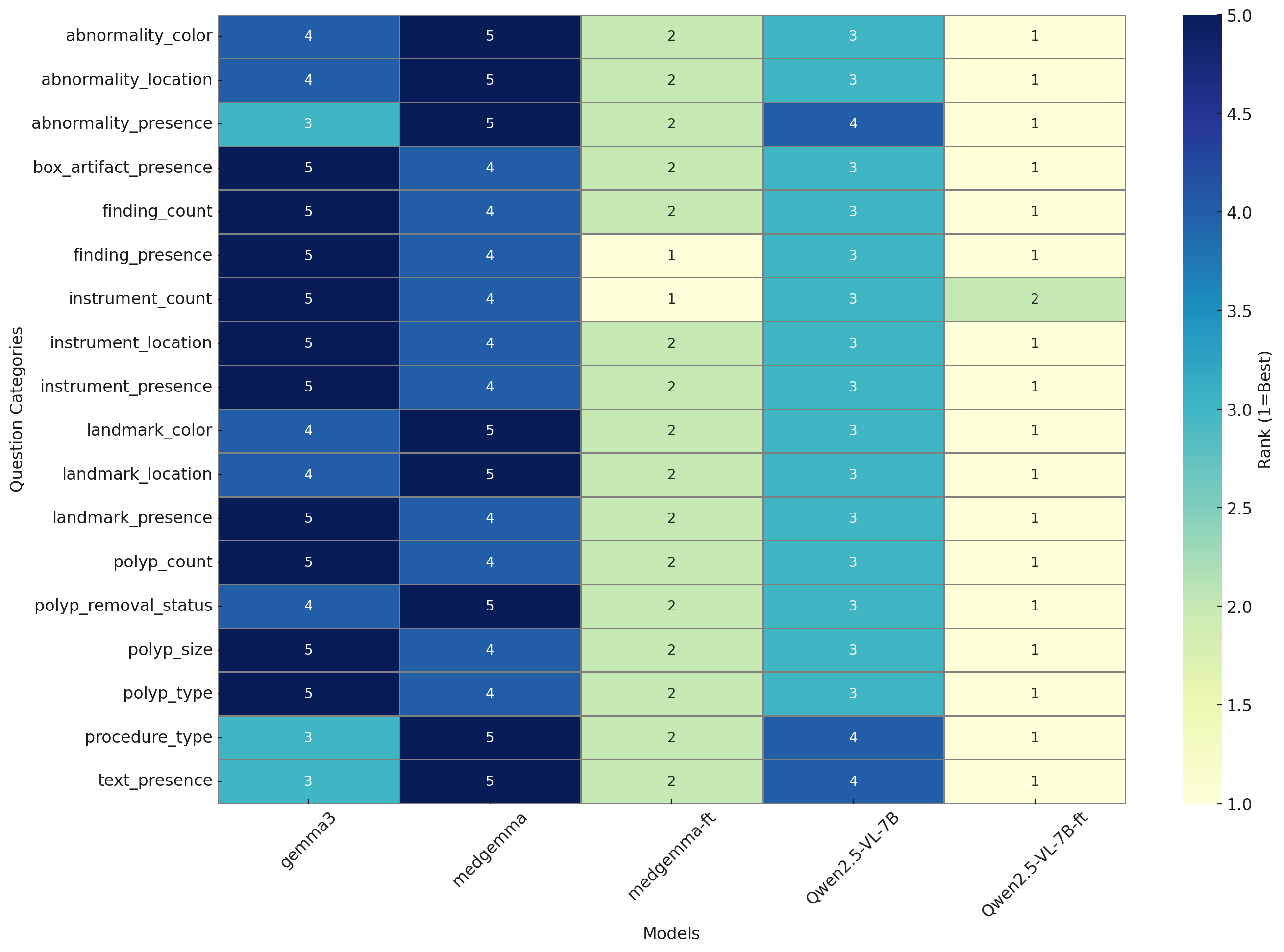}
    \caption{Rank-normalized heatmap illustrating comparative performance rankings (1 = best, 5 = worst) of the models across Kvasir-VQA categories. Qwen2.5-VL-7B-FT consistently ranks first across most categories.}
    \label{fig:model-heatmap-ranks}
\end{figure}

\begin{figure}[htpb]
\centering
    \includegraphics[width=\linewidth]{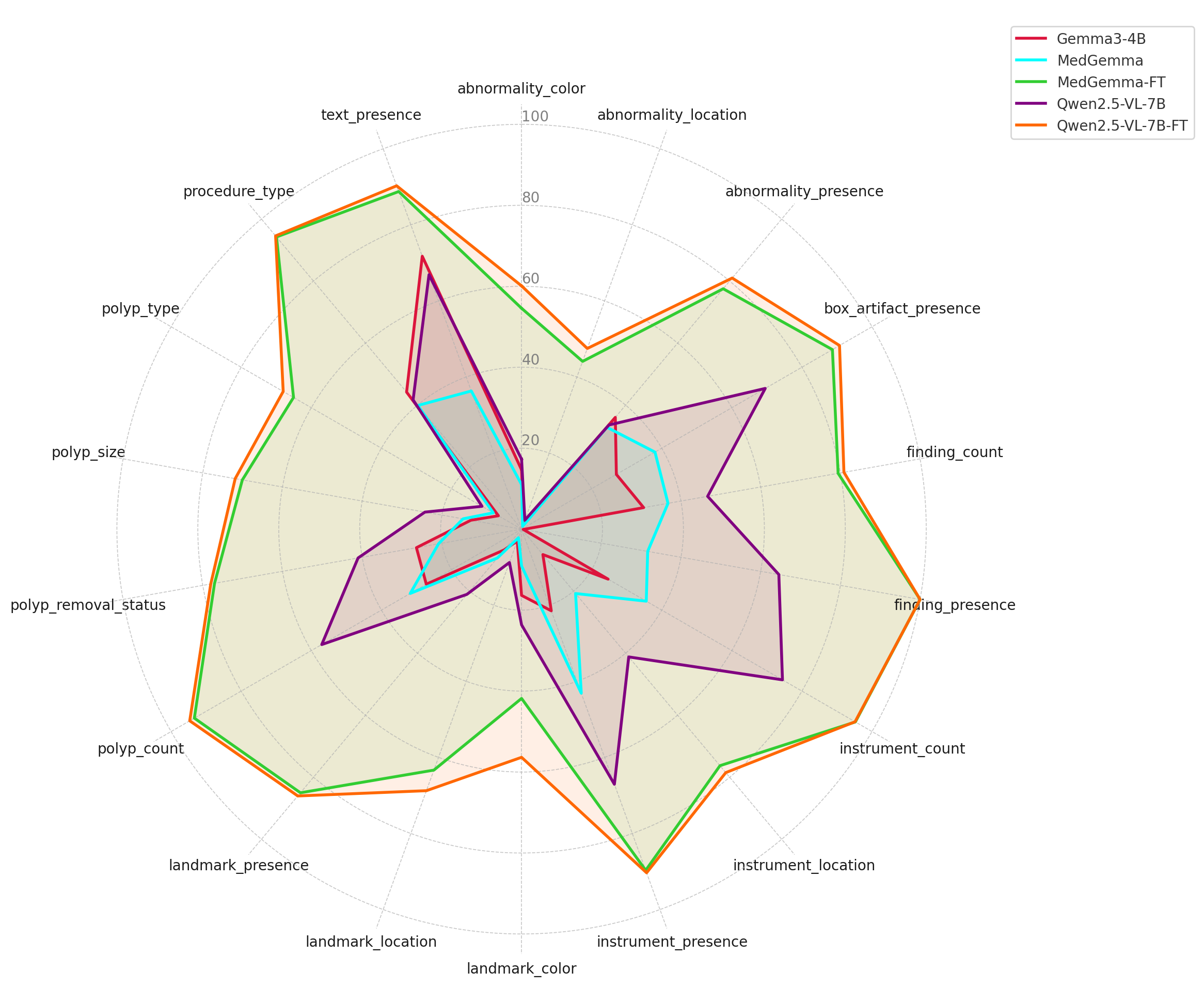}
    \caption{Radar plot showing absolute performance scores of five models (Gemma3-4B, MedGemma, MedGemma-FT, Qwen2.5-VL-7B, and Qwen2.5-VL-7B-FT) across various question categories. Higher values indicate better performance.}
    \label{fig:model-radar-performance}
\end{figure}

\begin{itemize}
    \item Radar Charts (Figure \ref{fig:model-radar-performance}) illustrated the absolute performance scores of the five models (Gemma3-4B, MedGemma, MedGemma-FT, Qwen2.5-VL-7B, and Qwen2.5-VL-7B-FT) across these categories. Higher values consistently indicated better performance. MedGemma-FT and Qwen2.5-VL-7B-FT demonstrated notable improvements over their base models in several clinical domains.
    \item Rank-Normalized Heatmaps (Figure \ref{fig:model-heatmap-ranks}) further clarified the comparative performance. This visualization assigned ranks (1 = best, 5 = worst) to models within each category. Qwen2.5-VL-7B-FT consistently ranked first across most categories, demonstrating its superior ability to answer questions across a wide range of clinical scenarios after fine-tuning. MedGemma-FT also showed strong relative performance compared to its base version.
\end{itemize}

\textbf{Complexity-Based Performance} \\

\begin{figure}[htpb]
    \centering
    \subfigure[Complexity Level 1]{
        \includegraphics[width=0.31\linewidth]{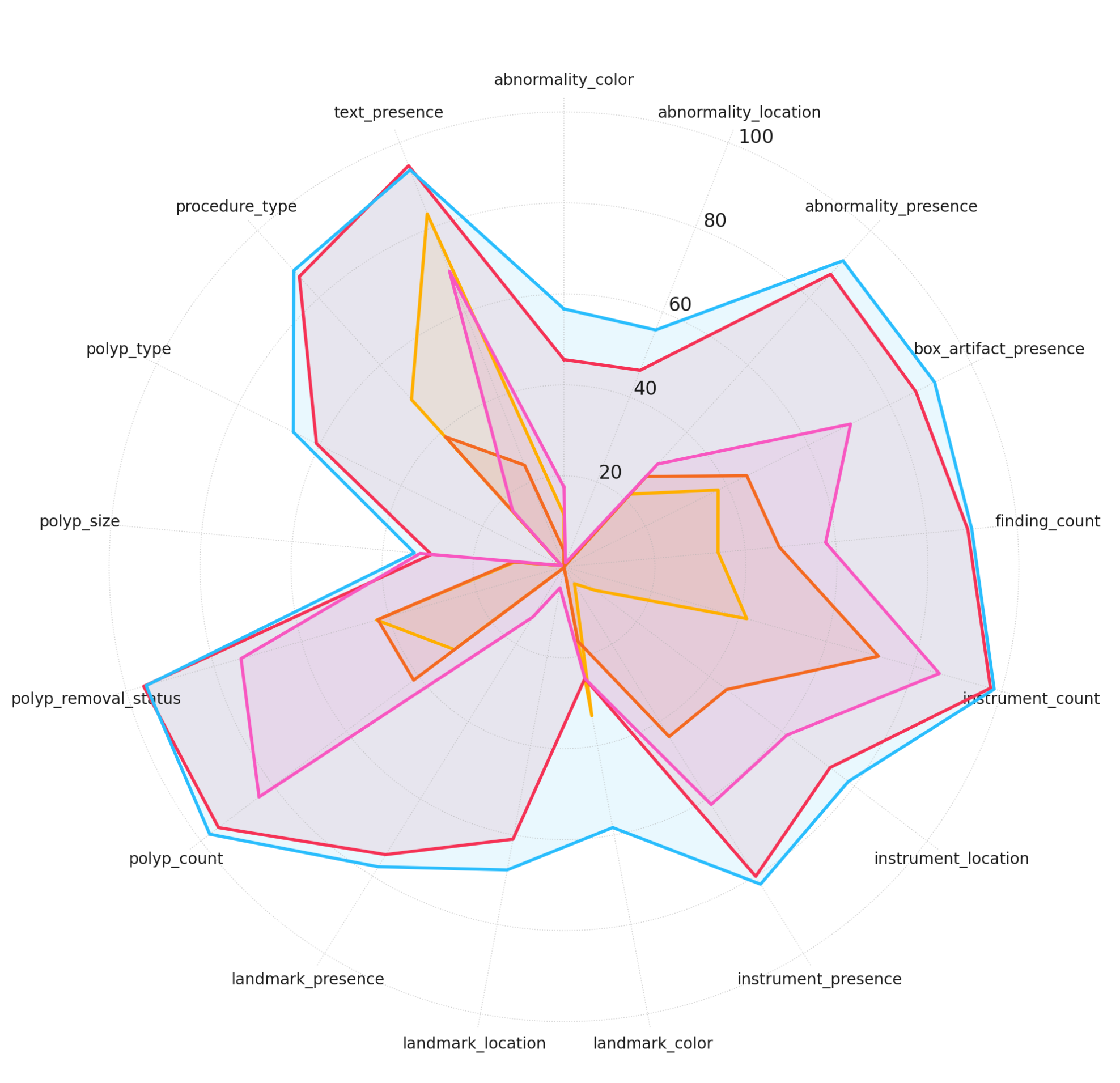}
    \label{fig:radar-complex-1}
    }
    \hfill
    \subfigure[Complexity Level 2]{
        \includegraphics[width=0.31\linewidth]{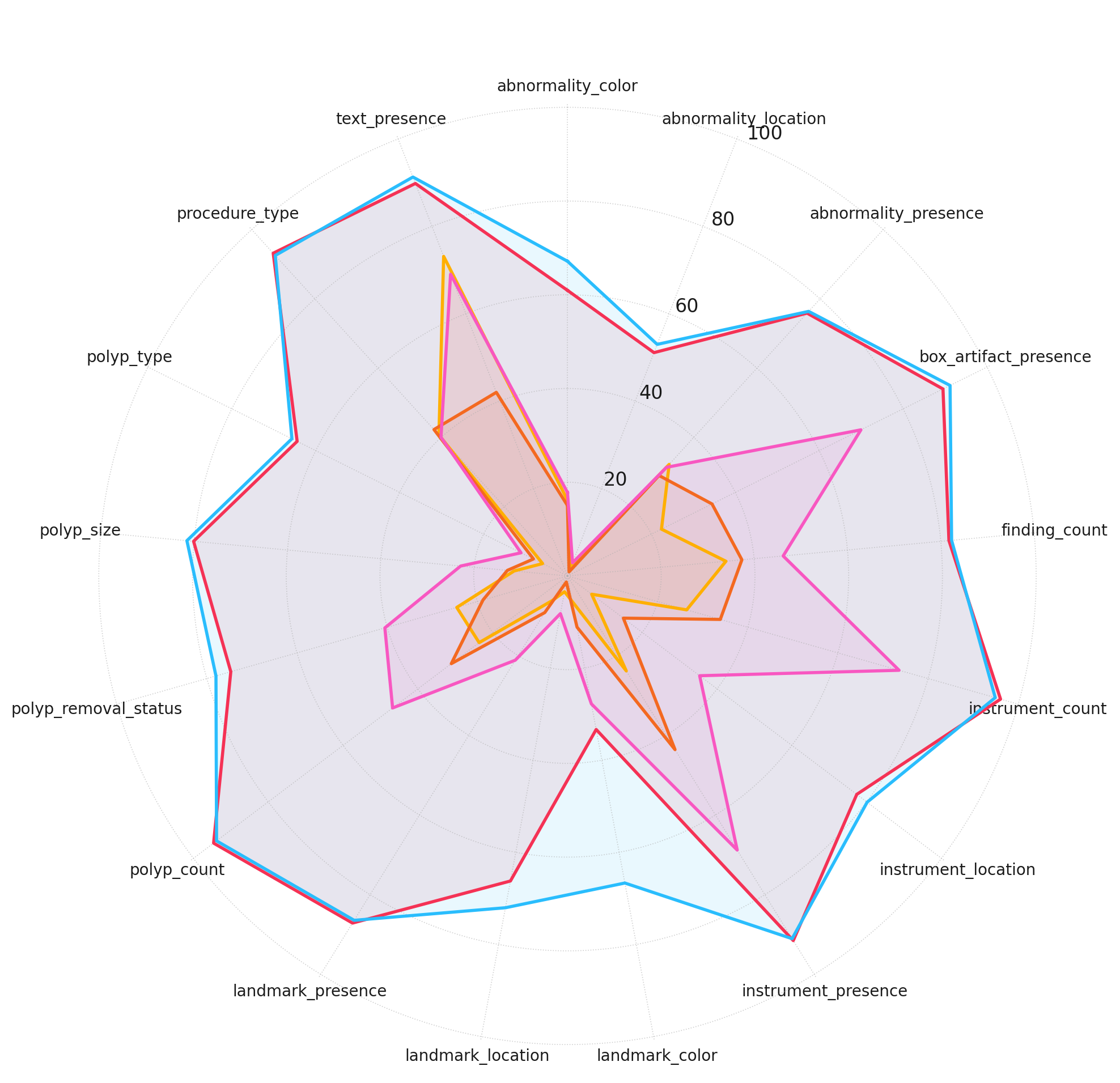}
    \label{fig:radar-complex-2}
    }
    \hfill
    \subfigure[Complexity Level 3]{
        \includegraphics[width=0.31\linewidth]{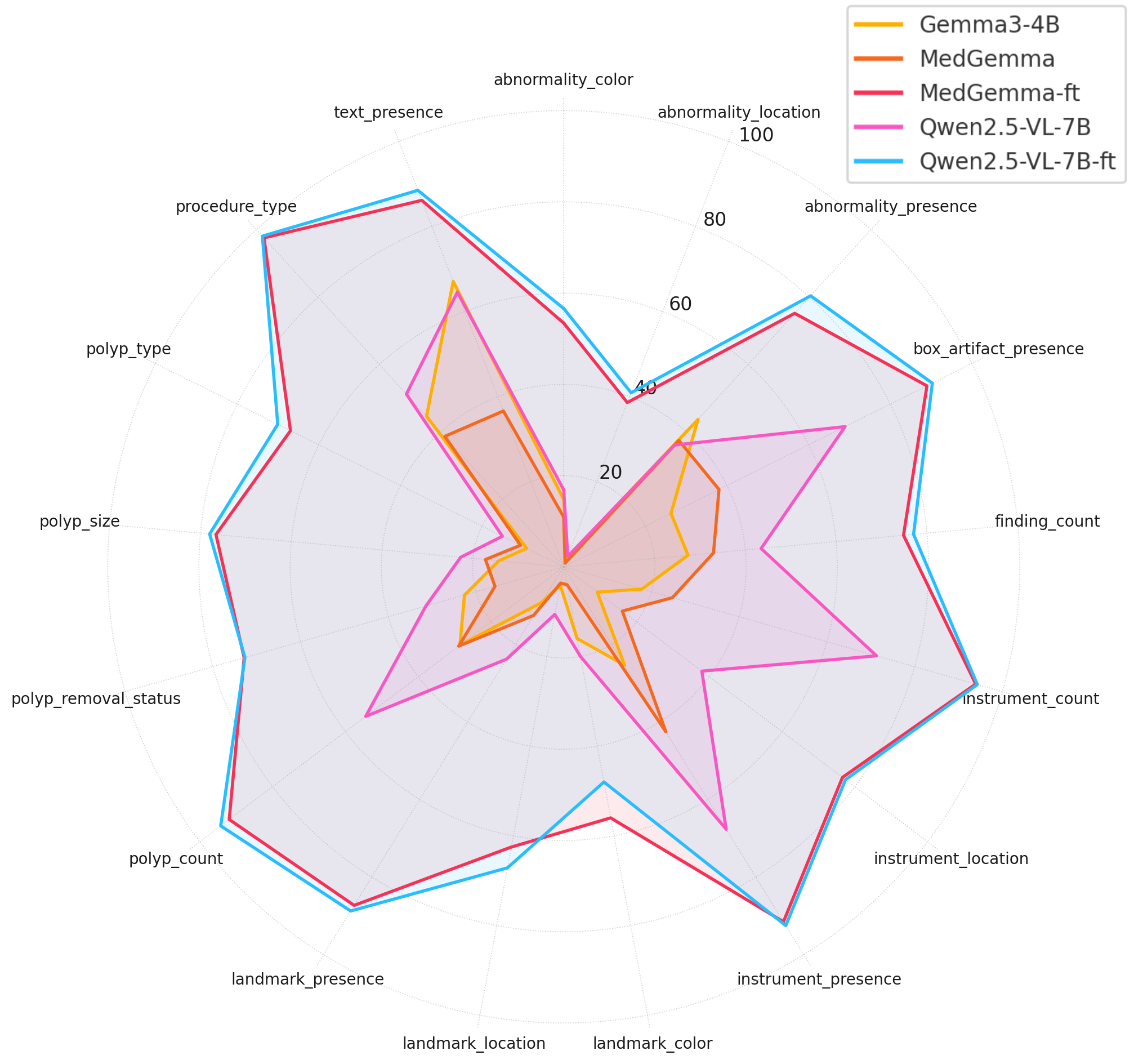}
    \label{fig:radar-complex-3}
    }
    \caption{Model performance across different complexity levels. Accuracy scores are plotted for each model across different question categories, grouped by reasoning complexity.}
    \label{fig:radar-complex-all}
\end{figure}

The analysis of performance across different reasoning complexity levels revealed how models handle increasing inferential demands (Figure \ref{fig:radar-complex-all}).

\begin{itemize}
    \item For Complexity Level 1 (Figure \ref{fig:radar-complex-1}), which involved direct factual recall, most fine-tuned models exhibited strong performance, indicating their proficiency in extracting straightforward information from images.
    \item At Complexity Level 2 (Figure \ref{fig:radar-complex-2}), requiring moderate reasoning and synthesis, the fine-tuned models, particularly Qwen2.5-VL-7B-FT, maintained a significant advantage, showcasing their ability to integrate information from multiple clinical cues.
    \item For Complexity Level 3 (Figure \ref{fig:radar-complex-3}), which demanded higher-order reasoning and abstraction across multiple clinical aspects, the fine-tuned models, especially Qwen2.5-VL-7B-FT, consistently outperformed their base counterparts. This indicated that fine-tuning significantly enhanced their capacity for complex clinical inference and cross-referencing.
\end{itemize}
Across all complexity levels, the fine-tuned models demonstrated increased accuracy, validating the effectiveness of the fine-tuning process in improving their ability to handle diverse linguistic and reasoning demands. Table \ref{tab:perf-across-complex} details model accuracy across all question
categories and complexity levels.

\begin{table}[htbp]
\tiny
\renewcommand{\arraystretch}{0.8}
\centering
\caption{Aspect-wise accuracy (\%) of different models across clinical question categories and reasoning complexity levels, computed using LLM-based adjudication. Each score reflects the proportion of correct responses per aspect (question class), where correctness is determined by a structured large language model evaluator assigning a binary score per aspect.}

\resizebox{\textwidth}{!}{%
\begin{tabular}{@{}lccccc@{}}

\textbf{Question Class} & \textbf{gemma3} & \textbf{medgemma} & \textbf{medgemma-ft} & \textbf{Qwen2.5-VL-7B} & \textbf{Qwen2.5-VL-7B-ft} \\
\toprule
 & \multicolumn{5}{c}{\textbf{Complexity Level 1}} \\

\midrule
abnormality\_color & 11.46 & 3.50 & 45.54 & 17.52 & 56.69 \\
abnormality\_location & 0.00 & 0.00 & 46.32 & 0.92 & 55.83 \\
abnormality\_presence & 21.60 & 26.85 & 87.04 & 30.56 & 91.05 \\
box\_artifact\_presence & 37.86 & 44.90 & 86.41 & 70.39 & 91.02 \\
finding\_count & 34.02 & 47.51 & 89.15 & 57.77 & 90.03 \\
finding\_presence & 0.42 & 31.65 & 100.00 & 64.56 & 100.00 \\
instrument\_count & 41.78 & 71.87 & 97.49 & 85.79 & 98.33 \\
instrument\_location & 8.61 & 44.81 & 73.29 & 61.42 & 78.34 \\
instrument\_presence & 4.39 & 43.92 & 80.07 & 61.49 & 82.09 \\
landmark\_color & 33.33 & 16.67 & 25.00 & 25.00 & 58.33 \\
landmark\_location & 0.00 & 0.00 & 60.94 & 4.72 & 67.81 \\
landmark\_presence & 0.45 & 0.45 & 74.44 & 13.00 & 77.58 \\
polyp\_count & 30.21 & 41.39 & 95.17 & 83.99 & 97.58 \\
polyp\_removal\_status & 42.82 & 42.54 & 96.06 & 73.80 & 95.49 \\
polyp\_size & 10.90 & 11.21 & 29.28 & 31.78 & 33.02 \\
polyp\_type & 0.60 & 0.60 & 60.78 & 0.60 & 66.47 \\
procedure\_type & 49.74 & 38.66 & 86.34 & 16.75 & 88.14 \\
text\_presence & 83.21 & 23.95 & 94.57 & 69.63 & 93.58 \\
\midrule
 & \multicolumn{5}{c}{\textbf{Complexity Level 2}} \\
\midrule
abnormality\_color & 16.18 & 15.03 & 60.95 & 17.81 & 67.16 \\
abnormality\_location & 1.89 & 0.95 & 51.10 & 3.00 & 53.00 \\
abnormality\_presence & 32.21 & 29.01 & 75.89 & 31.37 & 76.39 \\
box\_artifact\_presence & 22.41 & 34.46 & 89.51 & 69.95 & 91.19 \\
finding\_count & 33.95 & 37.38 & 81.74 & 46.22 & 82.31 \\
instrument\_count & 26.44 & 33.90 & 96.07 & 73.56 & 94.90 \\
instrument\_location & 6.47 & 14.95 & 77.35 & 35.34 & 80.12 \\
instrument\_presence & 23.84 & 43.59 & 91.52 & 68.80 & 91.04 \\
landmark\_color & 5.56 & 11.11 & 33.33 & 27.78 & 66.67 \\
landmark\_location & 3.43 & 1.32 & 66.23 & 8.18 & 72.03 \\
landmark\_presence & 6.21 & 9.20 & 87.13 & 21.15 & 86.44 \\
polyp\_count & 23.61 & 31.06 & 94.61 & 46.75 & 93.82 \\
polyp\_removal\_status & 24.56 & 18.73 & 74.68 & 40.51 & 77.97 \\
polyp\_size & 11.44 & 12.88 & 80.11 & 22.89 & 81.55 \\
polyp\_type & 5.93 & 8.11 & 64.43 & 11.08 & 65.68 \\
procedure\_type & 40.71 & 42.28 & 93.19 & 40.05 & 92.54 \\
text\_presence & 73.13 & 42.01 & 89.82 & 69.04 & 91.29 \\
\midrule
 & \multicolumn{5}{c}{\textbf{Complexity Level 3}} \\
\midrule
abnormality\_color & 14.74 & 10.90 & 53.42 & 16.88 & 56.62 \\
abnormality\_location & 2.13 & 0.85 & 38.62 & 2.45 & 40.91 \\
abnormality\_presence & 43.71 & 37.49 & 75.19 & 36.13 & 80.37 \\
box\_artifact\_presence & 26.28 & 38.04 & 88.99 & 68.93 & 90.31 \\
finding\_count & 27.38 & 32.99 & 74.84 & 43.47 & 77.05 \\
instrument\_count & 17.83 & 24.77 & 93.93 & 71.32 & 94.30 \\
instrument\_location & 9.24 & 16.11 & 76.61 & 37.96 & 77.44 \\
instrument\_presence & 25.35 & 42.56 & 91.53 & 67.68 & 92.51 \\
landmark\_color & 16.00 & 4.00 & 56.00 & 20.00 & 48.00 \\
landmark\_location & 4.11 & 3.64 & 62.50 & 10.62 & 67.14 \\
landmark\_presence & 9.24 & 12.54 & 87.30 & 23.81 & 88.75 \\
polyp\_count & 28.53 & 28.86 & 91.91 & 54.42 & 94.26 \\
polyp\_removal\_status & 22.62 & 15.67 & 72.86 & 31.42 & 72.69 \\
polyp\_size & 14.29 & 17.24 & 76.58 & 22.69 & 77.98 \\
polyp\_type & 9.08 & 10.60 & 66.94 & 15.02 & 70.06 \\
procedure\_type & 44.66 & 38.70 & 97.57 & 51.22 & 98.09 \\
text\_presence & 67.07 & 36.63 & 86.15 & 64.56 & 88.50 \\
\midrule
 & \multicolumn{5}{c}{\textbf{Overall }} \\
\midrule
abnormality\_color & 14.66 & 11.01 & 54.56 & 17.29 & 60.10 \\
abnormality\_location & 1.68 & 0.74 & 44.11 & 2.37 & 47.50 \\
abnormality\_presence & 36.04 & 32.84 & 77.52 & 33.59 & 80.98 \\
box\_artifact\_presence & 27.05 & 38.07 & 88.70 & 69.53 & 90.73 \\
finding\_count & 30.63 & 36.78 & 79.43 & 46.69 & 80.89 \\
finding\_presence & 0.42 & 31.65 & 100.00 & 64.56 & 100.00 \\
instrument\_count & 24.69 & 35.58 & 95.25 & 74.45 & 95.16 \\
instrument\_location & 8.19 & 20.76 & 76.28 & 41.20 & 78.51 \\
instrument\_presence & 21.47 & 43.13 & 89.69 & 67.07 & 90.34 \\
landmark\_color & 16.36 & 9.09 & 41.82 & 23.64 & 56.36 \\
landmark\_location & 3.14 & 2.25 & 63.34 & 8.77 & 68.76 \\
landmark\_presence & 6.69 & 9.32 & 85.02 & 21.04 & 86.04 \\
polyp\_count & 27.19 & 31.77 & 93.38 & 57.02 & 94.69 \\
polyp\_removal\_status & 26.38 & 20.84 & 77.04 & 41.03 & 77.99 \\
polyp\_size & 12.71 & 14.68 & 70.07 & 24.26 & 71.86 \\
polyp\_type & 6.61 & 8.09 & 65.07 & 11.29 & 68.02 \\
procedure\_type & 44.20 & 39.88 & 94.22 & 41.70 & 94.57 \\
text\_presence & 71.75 & 36.32 & 88.76 & 66.88 & 90.26 \\
\bottomrule
\end{tabular}
\label{tab:perf-across-complex}
}
\end{table}

\textbf{Robustness to Visual Perturbations through Augmentation-Based Fine-Tuning}

To assess the robustness of fine-tuned models under visual perturbations, we evaluated variants trained using augmented (transformed) images across both the original (normal) and transformed validation sets. Notably, the performance of these models—specifically Q-VL-ft-Trans-3000 and Q-VL-ft-Trans-4444—remained highly consistent across the two sets. The absolute differences across key metrics such as ROUGE-L, METEOR, BLEURT, and BERT-F1 were marginal, often within a range of 0.001–0.002. This stability indicates strong generalization capacity, even when evaluated on previously unseen perturbations.

In contrast, models trained exclusively on normal (unaugmented) images exhibited a modest decline in performance when evaluated on the transformed validation set. While the degradation was not severe, it was systematic—most notably reflected in a slight drop in ROUGE-L and BERT-F1 scores. Conversely, models trained on transformed data retained or slightly improved their performance when evaluated on the normal set, confirming that augmentation during training does not compromise performance on clean inputs but rather enhances generalizability.

\subsection{Discussion}

Our results offer a comprehensive view of how modern Vision-Language Models (VLMs) perform on complex clinical reasoning tasks. The findings reveal a clear narrative about the roles of fine-tuning, model scale, and architectural design, while also exposing the current frontiers of compositional reasoning.

\subsubsection{The Unifying Power of Fine-Tuning}

The most significant finding is the transformative impact of domain-specific fine-tuning. Across the board, the fine-tuned variants demonstrate a dramatic leap in performance, with \texttt{MedGemma-ft} and \texttt{Qwen2.5-VL-ft} achieving mean accuracies of 87\% and 90\%, respectively, compared to their base checkpoints' scores hovering around 30-45\%. This underscores a critical point: while large-scale pre-training provides essential foundational knowledge, it is insufficient for the nuanced demands of medical VQA. Fine-tuning on a high-quality, in-domain dataset like Kvasir-VQA-x1 is the dominant factor in unlocking clinical competency.

Interestingly, this intensive fine-tuning acts as a great equalizer. A purpose-built 4B parameter MedGemma, after tuning, performs nearly on par with a much larger 7B generalist Qwen model. This shows that the dataset's signal is incredibly strong, effectively aligning models of different scales and pre-training backgrounds to the specific task.

\subsubsection{Scale and Architecture: The Deciding Factors}

Despite the leveling effect of fine-tuning, the slight but consistent performance advantage of \texttt{Qwen2.5-VL-7B-ft} suggests that architectural superiority and scale become the deciding factors at the performance ceiling. We attribute Qwen's edge to two primary aspects:

\begin{enumerate}
    \item \textbf{Flexible Image Resolution:} Qwen's Vision Transformer (ViT) can process images at their native aspect ratio and dynamic resolutions. In contrast, MedGemma's SigLIP encoder uses a fixed size input. On a heterogeneous dataset like Kvasir, which aggregates images from various endoscopic systems, Qwen's flexibility likely preserves more contextual and fine-grained visual information, aiding in different tasks.
    \item \textbf{Hierarchical Vision Features:} Qwen's use of a hierarchical vision backbone (FPN-like features) might provide richer spatial cues, contributing to its stronger performance on localization-dependent tasks.
\end{enumerate}

\subsubsection{The Nuances of Reasoning Complexity: A Synthesis Sweet Spot}

A counter-intuitive yet critical finding is that for several categories, even the fine-tuned models achieve higher scores on Level 2 complexity questions than on the seemingly simpler Level 1 questions (e.g., for \texttt{Qwen2.5-VL-ft} in \textit{abnormality color}, L2 scores 67.16\% vs. L1's 56.69\%). This contradicts a simple monotonic difficulty scale and points to Level 2 as a "synthesis sweet spot," an optimal nexus of complexity and context that aligns perfectly with the models' fine-tuned capabilities. We attribute this to several factors grounded in our methodology:

\textbf{Contextual Richness over Atomic Recall:} The process of "Question Merging" and "Answer Naturalization" for Level 2 questions creates coherent prompts that are rich in context. For a model fine-tuned on this dataset, synthesizing information from two related clinical cues may be a more robust task than recalling a single, isolated, and potentially ambiguous atomic fact from a Level 1 query. The combined context in L2 questions provides more clues, reducing single-point failures. While Level 2 performance exceeds Level 1 in several categories, this may also be due to reduced ambiguity from merged prompts. For example, ‘What is the color of the abnormality?’ (L1) lacks specificity compared to ‘What is the color of the polyp and where is it located?’ (L2). Thus, improved scores may reflect clearer referents rather than higher reasoning alone.

\textbf{Optimized for Synthesis:} The explicit goal of the Kvasir-VQA-x1 dataset is to "promote reasoning beyond simple recall." The fine-tuning process therefore optimizes the models for exactly this kind of synthesis. Level 2, demanding "moderate reasoning and synthesis," represents the core challenge of the dataset, and the models' proficiency here reflects their successful adaptation to this core task.

\textbf{The High Cost of Higher-Order Reasoning (Level 3):} Conversely, the performance drop at Level 3 is pronounced and expected. Combining three distinct clinical facts exponentially increases the cognitive load. The primary driver of failure is likely error accumulation; a single error in perceiving one of the three components, or in their synthesis, results in a score of zero due to the strict "correctly and completely" criterion of the LLM adjudicator. This, combined with potential data sparsity for L3 examples and the challenge of true abstraction, firmly establishes Level 3 as a benchmark for future advances in multi-hop VQA.

\subsubsection{Effectiveness of Augmentation Strategies for Generalization}
The results highlight the effectiveness of incorporating visual augmentations during fine-tuning to improve model robustness. Models trained on transformed images demonstrated strong invariance to input perturbations, maintaining stable performance across both validation domains. This outcome reinforces the utility of data augmentation in clinical vision-language applications, where minor variations in endoscopic imagery are common due to differences in equipment, lighting, or procedural context.

By contrast, models trained solely on clean images showed limited resilience to such variations. Although their performance remained relatively high, the observed degradation on the transformed validation set suggests susceptibility to distributional shifts. Importantly, training with augmented data did not impair performance on the original images, suggesting no trade-off in fidelity.

These findings support the adoption of augmentation-informed training as a principled approach to enhance generalization. In clinical deployments where robustness and reliability are paramount, such strategies can be instrumental in reducing performance variance and ensuring consistent outputs across heterogeneous input conditions.

\subsubsection{Limitations and Future Directions}

While our study provides valuable insights, it has several limitations that open avenues for future work.
\subsubsection*{Limitations}
\begin{itemize}
    \item \textbf{Dataset Specificity:} Our analysis is confined to a single, albeit complex, sub-specialty of gastroenterology. The models' performance may not generalize to other medical domains like radiology or pathology without further fine-tuning.
    \item \textbf{Evaluation Protocol:} The LLM-based adjudicator, while powerful, is not infallible. The strict, binary scoring for complex questions may harshly penalize partially correct answers, particularly for Level 3, potentially underestimating a model's partial reasoning capabilities.
    \item \textbf{Persistent Error Modes:} Even the best models struggle with tasks requiring precise metric and spatial understanding (e.g., \textit{polyp size}, \textit{abnormality location}) and calibrated color perception (\textit{abnormality color}), indicating that current vision encoders or multi-modal feature projection techniques are not fully optimized for these fine-grained clinical assessments.

    \item \textbf{Homogeneity bias in LLM-as-a-Judge}:
    A key limitation of our evaluation protocol is the use of a Qwen-based LLM as the adjudicator, which introduces potential homogeneity bias. Since several evaluated models (e.g., Qwen2.5-VL-7B) share architectural lineage with the adjudicator, this overlap may result in self-enhancement bias, where congruent tokenization and latent representations lead to systematically favorable judgments.

\end{itemize}

\subsubsection*{Future Directions}
\begin{itemize}
    \item \textbf{Advanced Training Strategies:} Employ curriculum learning that leverages the "synthesis sweet spot," perhaps by starting with Level 2 questions to build a strong reasoning foundation before introducing simpler Level 1 recall tasks and more complex Level 3 abstraction challenges.
    \item \textbf{Explicit Spatial and Metric Supervision:} Enhance model training by incorporating auxiliary tasks, such as predicting bounding boxes for \textit{abnormality location} or adding segmentation masks to improve \textit{polyp size} estimation.
    \item \textbf{Data Augmentation:} Implement targeted augmentations, such as simulating variable lighting and white balance, to improve performance on color-dependent tasks.
    \item \textbf{Refined Evaluation:} Develop more nuanced evaluation protocols, such as ensemble adjudication or credit-based scoring for complex questions, to better handle cases of ``right answer, wrong wording'' and to provide credit for partially correct reasoning. To address the homogeneity bias in LLM-as-a-judge evaluation framework, future studies should incorporate adjudication using structurally distinct LLMs (e.g., Claude or Gemini) to ensure impartiality. Ideally, an ensemble of heterogeneous adjudicators should be employed to cross-validate scores and reduce the influence of architectural dependencies in automated evaluation.
\end{itemize}

\subsection{Conclusion}

In this paper, we have introduced Kvasir-VQA-x1, a comprehensive Visual Question Answering dataset designed to advance the development of multimodal AI systems in the field of gastrointestinal endoscopy. Our primary contribution is the creation of a large-scale resource that addresses key limitations of existing MedVQA datasets. By generating 159,549 contextually-rich as well as complex question-answer pairs, we have significantly increased the linguistic and reasoning diversity available for model training and evaluation.

A key innovation of our work is the structured approach to data creation. Through a pipeline assisted by large language models and validated by clinical experts, we have produced questions that require a deeper level of clinical understanding, moving beyond simple image recognition. The stratification of these questions by complexity, from single-fact retrieval to multi-step reasoning, provides a clear framework for assessing the inferential capabilities of AI models. Furthermore, the inclusion of visually augmented images allows for a thorough evaluation of model robustness, a critical factor for reliable deployment in real-world clinical environments where image quality can vary.

Our evaluation of leading vision-language models, MedGemma and Qwen2.5-VL, on Kvasir-VQA-x1 demonstrates the challenges posed by our dataset and highlights the performance gains achievable through fine-tuning. The detailed, category-based analysis reveals the specific strengths and weaknesses of current models in different areas of clinical reasoning.

For future work, the Kvasir-VQA-x1 dataset can serve as a foundational benchmark for developing the next generation of MedVQA systems. We anticipate that this resource will encourage research into more sophisticated models capable of nuanced, multi-step reasoning and greater resilience to visual perturbations. By making our dataset and evaluation scripts publicly available, we hope to foster a collaborative effort towards building more trustworthy and clinically-impactful AI in gastroenterology and other medical specialties.

\clearpage
\section{Usage Notes}
Kvasir-VQA-x1 is designed for flexible use in multimodal AI research, particularly in medical image understanding and clinical question answering. Researchers may use the dataset for:

\begin{itemize}
    \item \textbf{Training and evaluation} of vision-language models (VLMs), including instruction-tuned or generative systems.
    \item \textbf{Robustness analysis} through controlled perturbation-based testing using the provided augmentation scripts.
    \item \textbf{Curriculum learning or stratified benchmarking} by leveraging the provided complexity scores (Levels 1--3) for progressive evaluation of model reasoning.
    \item \textbf{Clinical interpretability research}, including hallucination detection, failure mode analysis, or human-in-the-loop evaluations.
\end{itemize}

Each QA instance can be associated with metadata in DataFrame/JSON format, enabling structured preprocessing, parsing, and filtering. Users are advised to refer to the accompanying documentation and sample code to ensure compatibility with model inputs and evaluation pipelines.

\vspace{1em}
\noindent
\textbf{Recommended Tools:} We encourage using the provided preprocessing scripts and evaluation toolkit, which support loading the dataset, augmenting images, and benchmarking model outputs.

\vspace{1em}
\noindent
\textbf{Licensing:} The dataset is released under \href{https://creativecommons.org/licenses/by-nc/4.0/}{Creative Commons Attribution Non Commercial 4.0} to foster widespread academic research and innovation in medical AI, subject to ethical use and citation.

\section{Code Availability}
The full codebase for dataset generation, image augmentation, training configurations, and evaluation scripts is publicly available at:

\url{https://github.com/simula/Kvasir-VQA-x1}

This repository includes:
\begin{itemize}
    \item Scripts for generating augmented images used in the transformed track.
    \item Preprocessing and JSON validation utilities.
    \item Sample training and evaluation workflows for fine-tuning VLMs.
    \item Baseline implementations for metrics and plotting tools (heatmaps, radar charts).
\end{itemize}

\section{Acknowledgements}
This work has benefited from the Experimental Infrastructure for Exploration of Exascale Computing (eX3), which is financially supported by the Research Council of Norway under contract 270053. We thank the medical experts who contributed to annotation and validation, as well as the SimulaMet and OsloMet infrastructure teams for supporting computational needs during large-scale fine-tuning.

\subsection*{Use of AI Disclosure}
Various AI/LLM tools were used to draft the structure and improve language clarity. All content has been carefully reviewed, verified, and finalized by the authors.

\scriptsize
\bibliography{references}

\begin{thebibliography}{54}
\providecommand{\natexlab}[1]{#1}
\providecommand{\url}[1]{\texttt{#1}}
\expandafter\ifx\csname urlstyle\endcsname\relax
  \providecommand{\doi}[1]{doi: #1}\else
  \providecommand{\doi}{doi: \begingroup \urlstyle{rm}\Url}\fi

\bibitem[Abbasian et~al.(2024)Abbasian, Khatibi, Azimi, Oniani, Shakeri Hossein~Abad, Thieme, Sriram, Yang, Wang, Lin, et~al.]{Abbasian2024Mar}
Mahyar Abbasian, Elahe Khatibi, Iman Azimi, David Oniani, Zahra Shakeri Hossein~Abad, Alexander Thieme, Ram Sriram, Zhongqi Yang, Yanshan Wang, Bryant Lin, et~al.
\newblock {Foundation metrics for evaluating effectiveness of healthcare conversations powered by generative AI}.
\newblock \emph{npj Digital Med.}, 7\penalty0 (82):\penalty0 1--14, March 2024.
\newblock ISSN 2398-6352.
\newblock \doi{10.1038/s41746-024-01074-z}.

\bibitem[Alayrac et~al.(2022)Alayrac, Donahue, Luc, Miech, Barr, Hasson, Lenc, Mensch, Millican, Reynolds, et~al.]{Flamingo}
Jean-Baptiste Alayrac, Jeff Donahue, Pauline Luc, Antoine Miech, Iain Barr, Yana Hasson, Karel Lenc, Arthur Mensch, Katie Millican, Malcolm Reynolds, et~al.
\newblock {Flamingo: a Visual Language Model for Few-Shot Learning}.
\newblock \emph{arXiv}, April 2022.
\newblock \doi{10.48550/arXiv.2204.14198}.

\bibitem[Ali et~al.(2019)Ali, Zhou, Bailey, Braden, East, Lu, and Rittscher]{Ali2019Apr}
Sharib Ali, Felix Zhou, Adam Bailey, Barbara Braden, James East, Xin Lu, and Jens Rittscher.
\newblock {A deep learning framework for quality assessment and restoration in video endoscopy}.
\newblock \emph{arXiv}, April 2019.
\newblock \doi{10.1016/j.media.2020.101900}.

\bibitem[Bai et~al.(2025)Bai, Chen, Liu, Wang, Ge, Song, Dang, Wang, Wang, Tang, Zhong, Zhu, Yang, Li, Wan, Wang, Ding, Fu, Xu, Ye, Zhang, Xie, Cheng, Zhang, Yang, Xu, and Lin]{Qwen2.5-VL}
Shuai Bai, Keqin Chen, Xuejing Liu, Jialin Wang, Wenbin Ge, Sibo Song, Kai Dang, Peng Wang, Shijie Wang, Jun Tang, Humen Zhong, Yuanzhi Zhu, Mingkun Yang, Zhaohai Li, Jianqiang Wan, Pengfei Wang, Wei Ding, Zheren Fu, Yiheng Xu, Jiabo Ye, Xi~Zhang, Tianbao Xie, Zesen Cheng, Hang Zhang, Zhibo Yang, Haiyang Xu, and Junyang Lin.
\newblock Qwen2.5-vl technical report.
\newblock \emph{arXiv preprint arXiv:2502.13923}, 2025.

\bibitem[Bazi et~al.(2023)Bazi, Rahhal, Bashmal, and Zuair]{Bazi2023Mar}
Yakoub Bazi, Mohamad Mahmoud~Al Rahhal, Laila Bashmal, and Mansour Zuair.
\newblock {Vision{\textendash}Language Model for Visual Question Answering in Medical Imagery}.
\newblock \emph{Bioengineering}, 10\penalty0 (3):\penalty0 380, March 2023.
\newblock ISSN 2306-5354.
\newblock \doi{10.3390/bioengineering10030380}.

\bibitem[Borgli et~al.(2020)Borgli, Thambawita, Smedsrud, Hicks, Jha, Eskeland, Randel, Pogorelov, Lux, Nguyen, Johansen, Griwodz, Stensland, Garcia-Ceja, Schmidt, Hammer, Riegler, Halvorsen, and de~Lange]{HyperKvasir}
Hanna Borgli, Vajira Thambawita, Pia~H. Smedsrud, Steven Hicks, Debesh Jha, Sigrun~L. Eskeland, Kristin~Ranheim Randel, Konstantin Pogorelov, Mathias Lux, Duc Tien~Dang Nguyen, Dag Johansen, Carsten Griwodz, H{\aa}kon~K. Stensland, Enrique Garcia-Ceja, Peter~T. Schmidt, Hugo~L. Hammer, Michael~A. Riegler, P{\aa}l Halvorsen, and Thomas de~Lange.
\newblock {HyperKvasir, a comprehensive multi-class image and video dataset for gastrointestinal endoscopy}.
\newblock \emph{Sci. Data}, 7\penalty0 (283):\penalty0 1--14, August 2020.
\newblock ISSN 2052-4463.
\newblock \doi{10.1038/s41597-020-00622-y}.

\bibitem[Buslaev et~al.(2018)Buslaev, Parinov, Khvedchenya, Iglovikov, and Kalinin]{Albumentations}
Alexander Buslaev, Alex Parinov, Eugene Khvedchenya, Vladimir~I. Iglovikov, and Alexandr~A. Kalinin.
\newblock {Albumentations: fast and flexible image augmentations}.
\newblock \emph{arXiv}, September 2018.
\newblock \doi{10.3390/info11020125}.

\bibitem[Chen et~al.(2020)Chen, Kornblith, Norouzi, and Hinton]{SimCLR}
Ting Chen, Simon Kornblith, Mohammad Norouzi, and Geoffrey Hinton.
\newblock {A Simple Framework for Contrastive Learning of Visual Representations}.
\newblock \emph{arXiv}, February 2020.
\newblock \doi{10.48550/arXiv.2002.05709}.

\bibitem[Chen et~al.(2024)Chen, Lai, Ruan, Chen, Liu, and Liu]{Chen2024Oct}
Xupeng Chen, Zhixin Lai, Kangrui Ruan, Shichu Chen, Jiaxiang Liu, and Zuozhu Liu.
\newblock {R-LLaVA: Improving Med-VQA Understanding through Visual Region of Interest}.
\newblock \emph{arXiv}, October 2024.
\newblock \doi{10.48550/arXiv.2410.20327}.

\bibitem[Dong et~al.(2025)Dong, Shen, Han, Tan, Wu, and Xu]{Dong2025Mar}
Wenjie Dong, Shuhao Shen, Yuqiang Han, Tao Tan, Jian Wu, and Hongxia Xu.
\newblock {Generative Models in Medical Visual Question Answering: A Survey}.
\newblock \emph{Appl. Sci.}, 15\penalty0 (6):\penalty0 2983, March 2025.
\newblock ISSN 2076-3417.
\newblock \doi{10.3390/app15062983}.

\bibitem[Gao et~al.(2025)Gao, Hu, Yin, Ruan, Pu, and Wan]{Gao2025}
Mingqi Gao, Xinyu Hu, Xunjian Yin, Jie Ruan, Xiao Pu, and Xiaojun Wan.
\newblock {LLM-based NLG Evaluation: Current Status and Challenges}.
\newblock \emph{Computational Linguistics}, pages 1--27, 2025.
\newblock \doi{10.1162/coli_a_00561}.

\bibitem[Gautam et~al.(2024)Gautam, Storås, Midoglu, Hicks, Thambawita, Halvorsen, and Riegler]{kvasirvqa}
Sushant Gautam, Andrea Storås, Cise Midoglu, Steven~A. Hicks, Vajira Thambawita, Pål Halvorsen, and Michael~A. Riegler.
\newblock Kvasir-vqa: A text-image pair gi tract dataset.
\newblock In \emph{Proceedings of the First International Workshop on Vision-Language Models for Biomedical Applications (VLM4Bio '24)}, page 10 pages. ACM, 2024.
\newblock \doi{10.1145/3689096.3689458}.

\bibitem[Google(2025)]{MedGemma}
Google.
\newblock Medgemma hugging face, May 2025.
\newblock URL \url{https://huggingface.co/collections/google/medgemma-release-680aade845f90bec6a3f60c4}.
\newblock [Online; accessed 29. May 2025].

\bibitem[Gu et~al.(2024)Gu, Yang, Liu, and Cai]{Gu2024Apr}
Tiancheng Gu, Kaicheng Yang, Dongnan Liu, and Weidong Cai.
\newblock {LaPA: Latent Prompt Assist Model For Medical Visual Question Answering}.
\newblock \emph{arXiv}, April 2024.
\newblock \doi{10.48550/arXiv.2404.13039}.

\bibitem[Guo et~al.(2025)Guo, Zhao, Wang, Chen, Liu, and Zhou]{Guo2025Mar}
Erjian Guo, Zhen Zhao, Zicheng Wang, Tong Chen, Yunyi Liu, and Luping Zhou.
\newblock {DiN: Diffusion Model for Robust Medical VQA with Semantic Noisy Labels}.
\newblock \emph{arXiv}, March 2025.
\newblock \doi{10.48550/arXiv.2503.18536}.

\bibitem[Hartsock and Rasool(2024)]{Hartsock2024Nov}
Iryna Hartsock and Ghulam Rasool.
\newblock {Vision-language models for medical report generation and visual question answering: a review}.
\newblock \emph{Front. Artif. Intell.}, 7:\penalty0 1430984, November 2024.
\newblock ISSN 2624-8212.
\newblock \doi{10.3389/frai.2024.1430984}.

\bibitem[He et~al.(2021)He, Gao, and Chen]{DeBERTaV3}
Pengcheng He, Jianfeng Gao, and Weizhu Chen.
\newblock {DeBERTaV3: Improving DeBERTa using ELECTRA-Style Pre-Training with Gradient-Disentangled Embedding Sharing}.
\newblock \emph{arXiv}, November 2021.
\newblock \doi{10.48550/arXiv.2111.09543}.

\bibitem[Hicks et~al.(2023{\natexlab{a}})Hicks, Stor{\aa}s, Halvorsen, de~Lange, Riegler, and Thambawita]{MedVQA-GI2023}
Steven Hicks, Andrea~M Stor{\aa}s, P{\aa}l Halvorsen, Thomas de~Lange, Michael Riegler, and Vajira Thambawita.
\newblock Overview of imageclefmedical 2023-medical visual question answering for gastrointestinal tract.
\newblock In \emph{CLEF (Working Notes)}, pages 1316--1327, 2023{\natexlab{a}}.

\bibitem[Hicks et~al.(2023{\natexlab{b}})Hicks, Stor{\aa}s, Halvorsen, de~Lange, Riegler, and Thambawita]{hicks2023overview}
Steven Hicks, Andrea~M Stor{\aa}s, P{\aa}l Halvorsen, Thomas de~Lange, Michael Riegler, and Vajira Thambawita.
\newblock Overview of imageclefmedical 2023-medical visual question answering for gastrointestinal tract.
\newblock In \emph{CLEF (Working Notes)}, pages 1316--1327, 2023{\natexlab{b}}.

\bibitem[Hu et~al.(2021)Hu, Shen, Wallis, Allen-Zhu, Li, Wang, Wang, and Chen]{lora}
Edward~J. Hu, Yelong Shen, Phillip Wallis, Zeyuan Allen-Zhu, Yuanzhi Li, Shean Wang, Lu~Wang, and Weizhu Chen.
\newblock {LoRA: Low-Rank Adaptation of Large Language Models}.
\newblock \emph{arXiv}, June 2021.
\newblock \doi{10.48550/arXiv.2106.09685}.

\bibitem[Hu et~al.(2023)Hu, Gu, An, Zhang, Liu, Kobayashi, Harada, Summers, and Zhu]{Hu2023Jul}
Xinyue Hu, Lin Gu, Qiyuan An, Mengliang Zhang, Liangchen Liu, Kazuma Kobayashi, Tatsuya Harada, Ronald~M. Summers, and Yingying Zhu.
\newblock {Expert Knowledge-Aware Image Difference Graph Representation Learning for Difference-Aware Medical Visual Question Answering}.
\newblock \emph{arXiv}, July 2023.
\newblock \doi{10.1145/3580305.3599819}.

\bibitem[Islam et~al.(2024)Islam, Hafiz, Jim, Kabir, and Mridha]{Islam2024Jun}
Tauhidul Islam, {\relax Md}.~Sadman Hafiz, Jamin~Rahman Jim, {\relax Md}.~Mohsin Kabir, and M.~F. Mridha.
\newblock {A systematic review of deep learning data augmentation in medical imaging: Recent advances and future research directions}.
\newblock \emph{Healthcare Analytics}, 5:\penalty0 100340, June 2024.
\newblock ISSN 2772-4425.
\newblock \doi{10.1016/j.health.2024.100340}.

\bibitem[Jha et~al.(2021)Jha, Ali, Emanuelsen, Hicks, Thambawita, Garcia-Ceja, Riegler, de~Lange, Schmidt, Johansen, Johansen, and Halvorsen]{KvasirInstrument}
Debesh Jha, Sharib Ali, Krister Emanuelsen, Steven~A. Hicks, Vajira Thambawita, Enrique Garcia-Ceja, Michael~A. Riegler, Thomas de~Lange, Peter~T. Schmidt, H{\aa}vard~D. Johansen, Dag Johansen, and P{\aa}l Halvorsen.
\newblock {Kvasir-Instrument: Diagnostic and Therapeutic Tool Segmentation Dataset in Gastrointestinal Endoscopy}.
\newblock In \emph{{MultiMedia Modeling}}, pages 218--229. Springer, Cham, Switzerland, January 2021.
\newblock ISBN 978-3-030-67835-7.
\newblock \doi{10.1007/978-3-030-67835-7_19}.

\bibitem[Lau et~al.(2018)Lau, Gayen, Ben~Abacha, and Demner-Fushman]{VQARAD}
Jason~J. Lau, Soumya Gayen, Asma Ben~Abacha, and Dina Demner-Fushman.
\newblock {A dataset of clinically generated visual questions and answers about radiology images}.
\newblock \emph{Sci. Data}, 5\penalty0 (180251):\penalty0 1--10, November 2018.
\newblock ISSN 2052-4463.
\newblock \doi{10.1038/sdata.2018.251}.

\bibitem[Lavie and Agarwal(2007)]{Meteor}
Alon Lavie and Abhaya Agarwal.
\newblock {Meteor: an automatic metric for MT evaluation with high levels of correlation with human judgments}.
\newblock In \emph{{DL Hosted proceedings}}, pages 228--231. Association for Computational Linguistics, June 2007.
\newblock \doi{10.5555/1626355.1626389}.

\bibitem[Li et~al.(2023)Li, Wong, Zhang, Usuyama, Liu, Yang, Naumann, Poon, and Gao]{Li2023Jun}
Chunyuan Li, Cliff Wong, Sheng Zhang, Naoto Usuyama, Haotian Liu, Jianwei Yang, Tristan Naumann, Hoifung Poon, and Jianfeng Gao.
\newblock {LLaVA-Med: Training a Large Language-and-Vision Assistant for Biomedicine in One Day}.
\newblock \emph{arXiv}, June 2023.
\newblock \doi{10.48550/arXiv.2306.00890}.

\bibitem[Lian et~al.(2024)Lian, Zhou, Yu, and Wang]{Lian2024Jan}
Chenyu Lian, Hong-Yu Zhou, Yizhou Yu, and Liansheng Wang.
\newblock {Less Could Be Better: Parameter-efficient Fine-tuning Advances Medical Vision Foundation Models}.
\newblock \emph{arXiv}, January 2024.
\newblock \doi{10.48550/arXiv.2401.12215}.

\bibitem[Liang et~al.(2024)Liang, Wang, Zhong, Wang, Li, Jia, and Wan]{Liang2024Sep}
Xiao Liang, Di~Wang, Haodi Zhong, Quan Wang, Ronghan Li, Rui Jia, and Bo~Wan.
\newblock {Candidate-Heuristic In-Context Learning: A new framework for enhancing medical visual question answering with LLMs}.
\newblock \emph{Information Processing {\&} Management}, 61\penalty0 (5):\penalty0 103805, September 2024.
\newblock ISSN 0306-4573.
\newblock \doi{10.1016/j.ipm.2024.103805}.

\bibitem[Lin(2004)]{Rouge}
Chin-Yew Lin.
\newblock Rouge: A package for automatic evaluation of summaries.
\newblock In \emph{Text summarization branches out}, pages 74--81, 2004.

\bibitem[Lin et~al.(2023)Lin, Zhang, Tao, Shi, Haffari, Wu, He, and Ge]{Lin2023Sep}
Zhihong Lin, Donghao Zhang, Qingyi Tao, Danli Shi, Gholamreza Haffari, Qi~Wu, Mingguang He, and Zongyuan Ge.
\newblock {Medical visual question answering: A survey}.
\newblock \emph{Artif. Intell. Med.}, 143:\penalty0 102611, September 2023.
\newblock ISSN 0933-3657.
\newblock \doi{10.1016/j.artmed.2023.102611}.

\bibitem[Liu et~al.(2021)Liu, Zhan, Xu, Ma, Yang, and Wu]{Slake}
Bo~Liu, Li-Ming Zhan, Li~Xu, Lin Ma, Yan Yang, and Xiao-Ming Wu.
\newblock {Slake: A Semantically-Labeled Knowledge-Enhanced Dataset For Medical Visual Question Answering}.
\newblock In \emph{{IEEE 18th International Symposium on Biomedical Imaging (ISBI)}}, pages 13--16. IEEE, 2021.
\newblock \doi{10.1109/ISBI48211.2021.9434010}.

\bibitem[Liu et~al.(2023)Liu, Li, Wu, and Lee]{laava}
Haotian Liu, Chunyuan Li, Qingyang Wu, and Yong~Jae Lee.
\newblock {Visual Instruction Tuning}.
\newblock \emph{arXiv}, April 2023.
\newblock \doi{10.48550/arXiv.2304.08485}.

\bibitem[Mangrulkar et~al.(2022)Mangrulkar, Gugger, Debut, Belkada, Paul, and Bossan]{peft-paper}
Sourab Mangrulkar, Sylvain Gugger, Lysandre Debut, Younes Belkada, Sayak Paul, and Benjamin Bossan.
\newblock Peft: State-of-the-art parameter-efficient fine-tuning methods.
\newblock In \emph{Peft: State-of-the-art parameter-efficient fine-tuning methods}. 2022.

\bibitem[Moor et~al.(2023{\natexlab{a}})Moor, Huang, Wu, Yasunaga, Zakka, Dalmia, Reis, Rajpurkar, and Leskovec]{MedFlamingo}
Michael Moor, Qian Huang, Shirley Wu, Michihiro Yasunaga, Cyril Zakka, Yash Dalmia, Eduardo~Pontes Reis, Pranav Rajpurkar, and Jure Leskovec.
\newblock {Med-Flamingo: a Multimodal Medical Few-shot Learner}.
\newblock \emph{arXiv}, July 2023{\natexlab{a}}.
\newblock \doi{10.48550/arXiv.2307.15189}.

\bibitem[Moor et~al.(2023{\natexlab{b}})Moor, Huang, Wu, Yasunaga, Zakka, Dalmia, Reis, Rajpurkar, and Leskovec]{Moor2023Jul}
Michael Moor, Qian Huang, Shirley Wu, Michihiro Yasunaga, Cyril Zakka, Yash Dalmia, Eduardo~Pontes Reis, Pranav Rajpurkar, and Jure Leskovec.
\newblock {Med-Flamingo: a Multimodal Medical Few-shot Learner}.
\newblock \emph{arXiv}, July 2023{\natexlab{b}}.
\newblock \doi{10.48550/arXiv.2307.15189}.

\bibitem[OpenAI et~al.(2023)OpenAI, Achiam, Adler, Agarwal, Ahmad, Akkaya, Aleman, Almeida, Altenschmidt, Altman, et~al.]{gpt4}
OpenAI, Josh Achiam, Steven Adler, Sandhini Agarwal, Lama Ahmad, Ilge Akkaya, Florencia~Leoni Aleman, Diogo Almeida, Janko Altenschmidt, Sam Altman, et~al.
\newblock {GPT-4 Technical Report}.
\newblock \emph{arXiv}, March 2023.
\newblock \doi{10.48550/arXiv.2303.08774}.

\bibitem[Ostmeier et~al.(2024)Ostmeier, Xu, Chen, Varma, Blankemeier, Bluethgen, Michalson, Moseley, Langlotz, Chaudhari, et~al.]{GREEN}
Sophie Ostmeier, Justin Xu, Zhihong Chen, Maya Varma, Louis Blankemeier, Christian Bluethgen, Arne~Edward Michalson, Michael Moseley, Curtis Langlotz, Akshay~S. Chaudhari, et~al.
\newblock {GREEN: Generative Radiology Report Evaluation and Error Notation}.
\newblock \emph{arXiv}, May 2024.
\newblock \doi{10.18653/v1/2024.findings-emnlp.21}.

\bibitem[Papineni et~al.(2002)Papineni, Roukos, Ward, and Zhu]{BLEU}
Kishore Papineni, Salim Roukos, Todd Ward, and Wei-Jing Zhu.
\newblock {BLEU: a method for automatic evaluation of machine translation}.
\newblock In \emph{{DL Hosted proceedings}}, pages 311--318. Association for Computational Linguistics, July 2002.
\newblock \doi{10.3115/1073083.1073135}.

\bibitem[Popovi{\ifmmode\acute{c}\else\'{c}\fi}(2015)]{chrF}
Maja Popovi{\ifmmode\acute{c}\else\'{c}\fi}.
\newblock {chrF: character n-gram F-score for automatic MT evaluation}.
\newblock \emph{ACL Anthology}, pages 392--395, September 2015.
\newblock \doi{10.18653/v1/W15-3049}.

\bibitem[Rajbhandari et~al.(2019)Rajbhandari, Rasley, Ruwase, and He]{ZeRO}
Samyam Rajbhandari, Jeff Rasley, Olatunji Ruwase, and Yuxiong He.
\newblock {ZeRO: Memory Optimizations Toward Training Trillion Parameter Models}.
\newblock \emph{arXiv}, October 2019.
\newblock \doi{10.48550/arXiv.1910.02054}.

\bibitem[Safavi-Naini et~al.(2024)Safavi-Naini, Ali, Shahab, Shahhoseini, Savage, Rafiee, Samaan, Shabeeb, Ladak, Yang, et~al.]{SafaviNaini2024Aug}
Seyed Amir~Ahmad Safavi-Naini, Shuhaib Ali, Omer Shahab, Zahra Shahhoseini, Thomas Savage, Sara Rafiee, Jamil~S. Samaan, Reem~Al Shabeeb, Farah Ladak, Jamie~O. Yang, et~al.
\newblock {Vision-Language and Large Language Model Performance in Gastroenterology: GPT, Claude, Llama, Phi, Mistral, Gemma, and Quantized Models}.
\newblock \emph{arXiv}, August 2024.
\newblock \doi{10.48550/arXiv.2409.00084}.

\bibitem[Sellam et~al.(2020)Sellam, Das, and Parikh]{BLEURT}
Thibault Sellam, Dipanjan Das, and Ankur~P. Parikh.
\newblock {BLEURT: Learning Robust Metrics for Text Generation}.
\newblock \emph{arXiv}, April 2020.
\newblock \doi{10.48550/arXiv.2004.04696}.

\bibitem[Singhal et~al.(2025)Singhal, Tu, Gottweis, Sayres, Wulczyn, Amin, Hou, Clark, Pfohl, Cole-Lewis, et~al.]{Singhal2025Mar}
Karan Singhal, Tao Tu, Juraj Gottweis, Rory Sayres, Ellery Wulczyn, Mohamed Amin, Le~Hou, Kevin Clark, Stephen~R. Pfohl, Heather Cole-Lewis, et~al.
\newblock {Toward expert-level medical question answering with large language models}.
\newblock \emph{Nat. Med.}, 31\penalty0 (3):\penalty0 943--950, March 2025.
\newblock ISSN 1546-170X.
\newblock \doi{10.1038/s41591-024-03423-7}.

\bibitem[Team(2025)]{qwen3technicalreport}
Qwen Team.
\newblock Qwen3 technical report, 2025.
\newblock URL \url{https://arxiv.org/abs/2505.09388}.

\bibitem[Wang et~al.(2023)Wang, Sanders, Liu, Seang, Tran, Atanasov, Qiu, Tang, Car, Wang, et~al.]{Wang2023Dec}
Xiaofei Wang, Hayley~M. Sanders, Yuchen Liu, Kennarey Seang, Bach~Xuan Tran, Atanas~G. Atanasov, Yue Qiu, Shenglan Tang, Josip Car, Ya~Xing Wang, et~al.
\newblock {ChatGPT: promise and challenges for deployment in low- and middle-income countries}.
\newblock \emph{Lancet Regional Health {\textendash} Western Pacific}, 41, December 2023.
\newblock ISSN 2666-6065.
\newblock \doi{10.1016/j.lanwpc.2023.100905}.

\bibitem[Wilkinson et~al.(2016)Wilkinson, Dumontier, Aalbersberg, Appleton, Axton, Baak, Blomberg, Boiten, da~Silva~Santos, Bourne, et~al.]{FAIR}
Mark~D. Wilkinson, Michel Dumontier, IJsbrand~Jan Aalbersberg, Gabrielle Appleton, Myles Axton, Arie Baak, Niklas Blomberg, Jan-Willem Boiten, Luiz~Bonino da~Silva~Santos, Philip~E. Bourne, et~al.
\newblock {The FAIR Guiding Principles for scientific data management and stewardship}.
\newblock \emph{Sci. Data}, 3\penalty0 (160018):\penalty0 1--9, March 2016.
\newblock ISSN 2052-4463.
\newblock \doi{10.1038/sdata.2016.18}.

\bibitem[Wolf et~al.(2020)Wolf, Debut, Sanh, Chaumond, Delangue, Moi, Cistac, Rault, Louf, Funtowicz, Davison, Shleifer, von Platen, Ma, Jernite, Plu, Xu, Scao, Gugger, Drame, Lhoest, and Rush]{transformers}
Thomas Wolf, Lysandre Debut, Victor Sanh, Julien Chaumond, Clement Delangue, Anthony Moi, Pierric Cistac, Tim Rault, Rémi Louf, Morgan Funtowicz, Joe Davison, Sam Shleifer, Patrick von Platen, Clara Ma, Yacine Jernite, Julien Plu, Canwen Xu, Teven~Le Scao, Sylvain Gugger, Mariama Drame, Quentin Lhoest, and Alexander~M. Rush.
\newblock Transformers: State-of-the-art natural language processing.
\newblock In \emph{Proceedings of the 2020 Conference on Empirical Methods in Natural Language Processing: System Demonstrations}, pages 38--45, Online, October 2020. Association for Computational Linguistics.
\newblock URL \url{https://www.aclweb.org/anthology/2020.emnlp-demos.6}.

\bibitem[Xu et~al.(2023)Xu, Xie, Qin, Tao, and Wang]{PEFT}
Lingling Xu, Haoran Xie, Si-Zhao~Joe Qin, Xiaohui Tao, and Fu~Lee Wang.
\newblock {Parameter-Efficient Fine-Tuning Methods for Pretrained Language Models: A Critical Review and Assessment}.
\newblock \emph{arXiv}, December 2023.
\newblock \doi{10.48550/arXiv.2312.12148}.

\bibitem[Yu et~al.(2025{\natexlab{a}})Yu, Wang, Wu, Xie, and Zhou]{MedFrameQA}
Suhao Yu, Haojin Wang, Juncheng Wu, Cihang Xie, and Yuyin Zhou.
\newblock {MedFrameQA: A Multi-Image Medical VQA Benchmark for Clinical Reasoning}.
\newblock \emph{arXiv}, May 2025{\natexlab{a}}.
\newblock \doi{10.48550/arXiv.2505.16964}.

\bibitem[Yu et~al.(2025{\natexlab{b}})Yu, Tong, Yu, and Zhang]{Yu2025Apr}
Ting Yu, Zixuan Tong, Jun Yu, and Ke~Zhang.
\newblock {Fine-grained Adaptive Visual Prompt for Generative Medical Visual Question Answering}.
\newblock \emph{AAAI}, 39\penalty0 (9):\penalty0 9662--9670, April 2025{\natexlab{b}}.
\newblock ISSN 2374-3468.
\newblock \doi{10.1609/aaai.v39i9.33047}.

\bibitem[Zhan et~al.(2020)Zhan, Liu, Fan, Chen, and Wu]{Zhan2020Oct}
Li-Ming Zhan, Bo~Liu, Lu~Fan, Jiaxin Chen, and Xiao-Ming Wu.
\newblock {Medical Visual Question Answering via Conditional Reasoning}.
\newblock In \emph{{ACM Conferences}}, pages 2345--2354. Association for Computing Machinery, New York, NY, USA, October 2020.
\newblock \doi{10.1145/3394171.3413761}.

\bibitem[Zhang et~al.(2019)Zhang, Kishore, Wu, Weinberger, and Artzi]{BERTScore}
Tianyi Zhang, Varsha Kishore, Felix Wu, Kilian~Q. Weinberger, and Yoav Artzi.
\newblock {BERTScore: Evaluating Text Generation with BERT}.
\newblock \emph{arXiv}, April 2019.
\newblock \doi{10.48550/arXiv.1904.09675}.

\bibitem[Zhang et~al.(2023)Zhang, Wu, Zhao, Lin, Zhang, Wang, and Xie]{PMCVQA}
Xiaoman Zhang, Chaoyi Wu, Ziheng Zhao, Weixiong Lin, Ya~Zhang, Yanfeng Wang, and Weidi Xie.
\newblock {PMC-VQA: Visual Instruction Tuning for Medical Visual Question Answering}.
\newblock \emph{arXiv}, May 2023.
\newblock \doi{10.48550/arXiv.2305.10415}.

\bibitem[Zhao et~al.(2024)Zhao, Huang, Hu, Wang, Mao, Zhang, Jiang, Wu, Ai, Wang, Zhou, and Chen]{ms-swift}
Yuze Zhao, Jintao Huang, Jinghan Hu, Xingjun Wang, Yunlin Mao, Daoze Zhang, Zeyinzi Jiang, Zhikai Wu, Baole Ai, Ang Wang, Wenmeng Zhou, and Yingda Chen.
\newblock Swift:a scalable lightweight infrastructure for fine-tuning, 2024.
\newblock URL \url{https://arxiv.org/abs/2408.05517}.

\end{thebibliography}

\end{document}